\documentclass[11pt]{article}

\usepackage[margin=1in]{geometry}
\usepackage{algorithm}
\usepackage{algorithmic}
\usepackage{amsmath, amssymb, amsthm}
\usepackage{mathtools}
\DeclarePairedDelimiter{\abs}{\lvert}{\rvert}
\usepackage{hyperref}
\usepackage[nameinlink,capitalize,noabbrev]{cleveref}
\usepackage{graphicx}
\usepackage{natbib}
\usepackage{url}            
\usepackage{bm}
\newcommand{\R}{\mathbb{R}}
\usepackage{booktabs}       
\usepackage{mathrsfs}      
\usepackage{amsfonts}       
\usepackage{nicefrac}       
\usepackage{microtype}      
\usepackage{xcolor}         
\usepackage{mathtools}
\usepackage{algorithm}
\usepackage{dsfont}
\usepackage{bbm}
\usepackage{longtable}
\usepackage{multirow}

\theoremstyle{plain}
\newtheorem{theorem}{Theorem}[section]
\newtheorem{proposition}[theorem]{Proposition}
\newtheorem{lemma}[theorem]{Lemma}
\newtheorem{corollary}[theorem]{Corollary}
\theoremstyle{definition}
\newtheorem{definition}[theorem]{Definition}

\theoremstyle{remark}
\newtheorem{remark}[theorem]{Remark}
\newtheorem{example}[theorem]{Example}

\title{Tight Asymptotic Bounds on the Maximum Number of
Linear Regions of Piecewise-Linear Kolmogorov--Arnold networks}
\author{
Ye Su\\
Shenzhen Institutes of Advanced Technology\\
Chinese Academy of Sciences\\
Shenzhen 518055, China\\
\texttt{ye.su@siat.ac.cn}
\and
Yong Liu\thanks{Corresponding author}\\
Gaoling School of Artificial Intelligence\\
Renmin University of China\\
Beijing 100872, China\\
\texttt{liuyonggsai@ruc.edu.cn}
}
\date{}

\begin{document}

\maketitle

\begin{abstract}
Schoots et al.~\citep{schoots2025relating} (AISTATS 2025) established a generally loose upper bound on the number of linear regions of continuous piecewise-linear (CPWL) Kolmogorov--Arnold networks (KANs), without providing
a lower bound. We consider a $T$-layer network with width vector $\mathbf n=(n_0,\ldots,n_T)$, input dimension $n_0=d$, output width $n_T=1$, and at most $k\geq2$ affine segments per edge. Our matching upper and lower bounds determine the tight growth order $\Theta_{\mathbf n}(k^{\sum_{\ell=1}^{T}\rho_{\ell-1}})$ in $k$, where $\rho_{\ell-1}:=\min_{0\leq s\leq\ell-1}n_s$. When every hidden width is at least $d$, the tight order simplifies to $\Theta_{\mathbf n}(k^{dT})$. The upper bound of \citep{schoots2025relating} is $k^{\sum_{\ell=1}^{T}n_{\ell-1}n_\ell}$. Our upper bound improves it by a factor of $\Theta_{\mathbf n}\!\left(k^{\sum_{\ell=1}^{T}n_{\ell-1}n_\ell-\sum_{\ell=1}^{T}\rho_{\ell-1}} \right)$, with strict asymptotic improvement whenever a hidden layer has width greater than one. We further compare CPWL KANs with rectified linear unit (ReLU) networks using dense parameterizations with at most $\Lambda$ parameters per model and the same fixed nonlinear depth $T$. Counting trainable knot locations and optimizing the widths and $k$, we prove that the maximum region count of CPWL KANs divided by that of ReLU networks is $\Theta_{d,T}(\Lambda^{d\lfloor T/2\rfloor})$ as $\Lambda\to\infty$, yielding a polynomial advantage for KANs when $T\geq2$. These tight bounds characterize the attainable geometric complexity of CPWL KANs, providing a foundation for understanding their expressive power and guiding architectural design.
\end{abstract}

\section{Introduction}
\label{sec:introduction}

Kolmogorov--Arnold networks (KANs) place learnable univariate functions on edges and combine their outputs by summation~\cite{liu2025kan}. This design makes the complexity of edge functions an architectural resource alongside width and depth \citep{somvanshi2025survey,dutta2026first}. We study continuous piecewise-linear (CPWL) KANs, whose edge functions consist of finitely many affine segments \citep{schoots2025relating}. A knot marks a change of slope between adjacent segments. The resulting networks admit a geometric description through partitions of the input space.

The maximum number of linear regions is a standard measure of CPWL expressive capacity \cite{montufar2014number,serra2018bounding,goujon2024number}. A linear region is a maximal nonempty connected open set on which the network agrees with an affine map \citep{hanin2019complexity}. Its count measures how finely the network can divide its input space. Previous work has established region-count bounds for rectified linear unit (ReLU) and general CPWL networks, revealing the effects of composition and intermediate dimensions \citep{montufar2014number,serra2018bounding,hu2020analysis,xiong2020number,xiong2024number}. For CPWL KANs, a sharp analysis must additionally account for the number and placement of knots on individual edges.

Consider a scalar-valued CPWL KAN $f:\R^d\to\R$, where $d$ is the input dimension. Let $T$ denote the number of KAN layers and $\mathbf n=(n_0,\ldots,n_T)$ its width vector, where $n_\ell$ is the width of layer $\ell$, $n_0=d$, and $n_T=1$. Thus, the network has $T-1$ hidden layers. Suppose that every edge function has at most $k\geq2$ affine segments, and let $\mathcal R(f)$ denote the number of linear regions of $f$. Schoots et al.~\cite{schoots2025relating} established the upper bound
\[
    \mathcal R(f)\leq k^{\sum_{\ell=1}^{T}n_{\ell-1}n_\ell}.
\]
This bound is generally loose because it counts segment choices separately for every edge and ignores the geometric constraints linking those choices. Their region-count analysis also provides no lower bound, leaving the attainable asymptotic order undetermined.

Two geometric properties explain the gap. First, edges leaving the same source node depend on the same scalar coordinate, so their knots produce parallel cuts within each cell where the preceding network is affine. Second, narrow layers limit the number of independent directions available to subsequent layers. We express the latter constraint through the effective-dimension bound $\rho_{\ell-1}:=\min_{0\leq s\leq\ell-1}n_s$ before layer $\ell$. These properties motivate an analysis based on arrangements of parallel hyperplanes.

In this paper, we consider two knot parameterizations. In the edge-wise-knot model, every edge chooses its own knots. In the source-shared-knot model, all edges leaving a source node use knots from one common candidate set, while their affine coefficients remain edge-specific. The shared model is a structured subclass of the edge-wise model. Studying both determines whether independent knot placement is necessary to attain the optimal growth rate. Our main contributions are as follows.
\begin{itemize}
    \item
    \textbf{Matching upper and lower bounds for arbitrary architectures.}
    For $\diamond\in\{\mathrm{ind},\mathrm{sh}\}$, let $\mathcal R^\star_\diamond(\mathbf n,k)$ denote the maximum number of linear regions in the corresponding edge-wise-knot or source-shared-knot class. We determine the tight growth order in $k$ for arbitrary architectures, including those with hidden widths smaller than $d$:
    \[
        \mathcal R^\star_\diamond(\mathbf n,k)
        =
        \Theta_{\mathbf n}\!\left(
            k^{\sum_{\ell=1}^{T}\rho_{\ell-1}}
        \right).
    \]
    Here $\Theta$ denotes matching upper and lower asymptotic orders, with its subscript specifying the allowed dependence of the constants. A source-shared-knot construction combines folding functions with coordinate grouping and produces cells with distinct final gradients, ensuring that they belong to different linear regions. Thus, shared knots already attain the optimal exponent. When every hidden width is at least $d$, this exponent simplifies to $dT$.

    \item
    \textbf{Quantified improvement over the existing KAN bound.}
    We derive explicit geometric upper bounds for both knot parameterizations that never exceed the upper bound $k^{\sum_{\ell=1}^{T}n_{\ell-1}n_\ell}$ of ~\cite{schoots2025relating}. The ratio of their upper bound to either of ours has growth order
    \[
        \Theta_{\mathbf n}\!\left(
            k^{
                \sum_{\ell=1}^{T}n_{\ell-1}n_\ell
                -
                \sum_{\ell=1}^{T}\rho_{\ell-1}
            }
        \right).
    \]
    Our bounds have a strictly smaller exponent whenever at least one hidden layer has width greater than one.

    \item
    \textbf{KAN--ReLU comparison with at most $\Lambda$ parameters.}
    We compare edge-wise-knot CPWL KANs and ReLU networks, allowing each model at most $\Lambda$ parameters. We count $2k$ parameter slots per KAN edge and all ReLU weights and biases, including zero entries. With $d$ and $T$ fixed, we optimize the widths and $k$. For $T$ KAN layers and $T$ ReLU hidden layers followed by an affine scalar output, the maximum region counts have orders
    \[
        \Theta_{d,T}\!\left(\Lambda^{dT}\right)
        \quad\text{and}\quad
        \Theta_{d,T}\!\left(
            \Lambda^{d\lceil T/2\rceil}
        \right),
    \]
    respectively. Consequently, KANs have a larger asymptotic maximum region count for $T\geq2$ under the stated parameter counts and depth convention.
\end{itemize}

\section{Related Work}
\label{sec:related-work}

The number of linear regions is a standard geometric measure of the expressive capacity of CPWL neural networks. Mont\'ufar et al.~\cite{montufar2014number} analyzed the maximum number of linear regions of ReLU and Maxout networks and showed through constructive lower bounds that this number can grow exponentially with depth. Serra, Tjandraatmadja, and Ramalingam~\cite{serra2018bounding} derived tighter upper and lower bounds for the maximum number of linear regions of deep ReLU networks and obtained an exact result for one-dimensional inputs under suitable width conditions. Hanin and Rolnick~\cite{hanin2019complexity} studied the expected, rather than maximum, number of linear regions along one-dimensional input subspaces and showed that it grows only linearly with the total number of neurons at random initialization. Hu et al.~\cite{hu2020analysis} analyzed networks with general piecewise-linear activations, obtaining the exact maximum number of regions for a single layer and upper and lower bounds for multilayer networks. Xiong et al.~\cite{xiong2020number} derived the maximum and average numbers of linear regions for one-layer ReLU convolutional neural networks (CNNs), together with bounds for deep CNNs. Mont\'ufar, Ren, and Zhang~\cite{montufar2022sharp} related the linear regions of single-layer Maxout networks to upper vertices of Minkowski sums and derived sharp single-layer upper bounds and asymptotically sharp multilayer upper bounds. Chen, Wang, and Xiong~\cite{chen2023lower} derived a sharp upper bound for the maximum number of linear regions of one-layer graph convolutional networks and upper and lower bounds for multilayer networks. Goujon et al.~\cite{goujon2024number} developed a general geometric framework for counting the maximum number of regions generated by CPWL networks, based on common refinements, sums, vectorizations, and compositions of convex partitions; their framework provides upper and lower bounds for architectures with univariate or multivariate CPWL activations, including deep-spline networks, and is therefore closely related to the piecewise-linear edge functions considered in this work. Xiong et al.~\cite{xiong2024number} extended the analysis of CNNs to general piecewise-linear activations and derived maximum-region bounds that account jointly for convolutional structure, activation complexity, width, and depth.
Pham and Garg~\cite{pham2024what} used tropical geometry to derive upper and lower bounds on the number of linear regions of ReLU message-passing graph neural networks. Bakaev et al.~\cite{bakaev2026better} used polyhedral subdivisions of simplices to improve the sufficient depth for exactly representing arbitrary CPWL functions with ReLU networks. Su et al.~\cite{su2026sparsity} used tropical geometry to analyze the maximum number of regions induced by Top-$k$ mixture-of-experts (MoE) routing, where $k$ is the number of experts selected for each input, and related the routing partition to the normal fan of a hypersimplex. Su and Liu~\cite{su2026expressivity} used tropical geometry to analyze the maximum number of full-dimensional affine regions of conditioned single-query Transformers in the zero-temperature hard-routing regime; with $N$ denoting sequence length, $H$ the number of attention heads, $d_{\mathrm{model}}$ the embedding dimension, and $D$ the number of Transformer blocks, they established matching upper and lower bounds of order $N^{\min\{H,d_{\mathrm{model}}-1\}D}$ as $N$ tends to infinity under their stated architectural conditions. Most closely related to the present work, Schoots et al.~\cite{schoots2025relating} studied the polyhedral decomposition of CPWL KANs. For a scalar-valued network $f:\R^d\to\R$ with $T$ layers, width vector $\mathbf n=(n_0,\ldots,n_T)$, input width $n_0=d$, output width $n_T=1$, and at most $k\geq2$ affine segments on every edge function, their decomposition bound implies
\[
    \mathcal R(f)\leq k^{\sum_{\ell=1}^{T}n_{\ell-1}n_\ell},
\]
where $\mathcal R(f)$ denotes the number of linear regions of $f$.
This edgewise product bound is generally loose. For a layer with input width $p$ and output width $q$, it counts up to $k^{pq}$ segment combinations. However, all outgoing edge functions from the same source depend on one scalar coordinate, so their knots generate
parallel cuts. The resulting refinement therefore contains at most
$\bigl[q(k-1)+1\bigr]^p$ full-dimensional cells. Across layers, the product bound also ignores reductions in effective dimension caused by narrow intermediate layers. Moreover, Schoots et al.~\cite{schoots2025relating} provided no lower bound on the maximum region count. Their result therefore leaves the attainable asymptotic dependence of the maximum region count on $k$ undetermined.

\section{Preliminaries}
\label{sec:preliminaries}

In this section, we introduce the mathematical tools and piecewise-linear KAN models used throughout the paper. We review CPWL maps, linear regions, polyhedral partitions, and parallel hyperplane arrangements, and then define the KAN architecture and the two knot parameterizations analyzed in the following sections. All Complete proofs are provided in Appendix~\ref{app:proofs}.

\subsection{Mathematical Tools}
\label{subsec:mathematical-tools}

Throughout the paper, let $\mathbb N:=\{1,2,\ldots\}$ and
$[m]:=\{1,\ldots,m\}$ for $m\in\mathbb N$. For a finite set $S$,
let $\abs{S}$ denote its cardinality. Let $p,q\in\mathbb N$ denote the input and output dimensions of the map under consideration, respectively. For a network layer, these are its input and output widths. An affine map from $\R^p$ to $\R^q$ has the form $x\mapsto Ax+b$, where $x\in\R^p$, $A\in\R^{q\times p}$, and $b\in\R^q$.

\begin{definition}[Polyhedral partition]
A convex polyhedron in $\R^p$ is a set $P=\{x\in\R^p:Cx\leq v\}$, where
$s\in\mathbb N$ is the number of linear inequalities, $C\in\R^{s\times p}$ is their coefficient matrix, and $v\in\R^s$ is constant vector. The inequality is interpreted componentwise. A finite family $\mathcal P=\{P_1,\ldots,P_M\}$ of convex polyhedra, with $M\in\mathbb N$, is a full-dimensional polyhedral partition of $\R^p$ if every $P_a$ has nonempty interior, $\bigcup_{a=1}^{M}P_a=\R^p$, and $\operatorname{int}(P_a)\cap\operatorname{int}(P_b)=\varnothing$ for all distinct $a,b\in[M]$. Its members are called cells, and their interiors are called open cells~\cite{goujon2024number}.
\end{definition}

Polyhedral partitions describe the cells on which a piecewise-linear network applies an affine formula.

\begin{definition}[Continuous piecewise-linear map]
A map $g:\R^p\to\R^q$ is continuous piecewise-linear, abbreviated as CPWL, if it is continuous and admits a full-dimensional polyhedral partition $\mathcal P=\{P_1,\ldots,P_M\}$ such that $g(x)=A_ax+b_a$ on every $P_a$, for some $A_a\in\R^{q\times p}$ and $b_a\in\R^q$
\cite{goujon2024number}.
\end{definition}

A partition may subdivide a set on which the same affine formula applies. We therefore count maximal connected affine regions.

\begin{definition}[Linear region]
A linear region of a CPWL map $g:\R^p\to\R^q$ is an inclusion-maximal nonempty connected open set $U\subseteq\R^p$ on which $g$ agrees with one affine map. The number of these regions is denoted by $\mathcal R(g)$
\cite{montufar2014number,serra2018bounding}.
\end{definition}

\begin{example}
The function $g(x)=\abs{x}$ has two linear regions, $(-\infty,0)$ and $(0,\infty)$, so $\mathcal R(g)=2$. Further subdividing either interval does not change this count.
\end{example}

The region-counting argument uses arrangements of hyperplanes. For KANs, knots associated with the same source coordinate produce parallel hyperplanes.

\begin{definition}[Hyperplane arrangement]
Let $r\in\mathbb N$ denote the dimension of the ambient space. For a nonzero normal vector $u\in\R^r$ and a scalar offset $\beta\in\R$ specifying the hyperplane's position, the set $H(u,\beta):=\{x\in\R^r:u^\top x=\beta\}$ is an affine hyperplane. A finite collection $\mathcal H$ of such hyperplanes is called a hyperplane arrangement. A finite collection $\mathcal H$ of such hyperplanes is a hyperplane arrangement. Its regions are the connected components of $\R^r\setminus\bigcup_{H\in\mathcal H}H$. A parallel family is a collection of mutually parallel hyperplanes. An arrangement is in general position if the normals of every collection of at most $r$ hyperplanes are linearly independent and no $r+1$ hyperplanes have a common point
\cite{zaslavsky1975facing}.
\end{definition}

The following standard bound accounts for parallel families
\cite{zaslavsky1975facing,goujon2024number}.

\begin{proposition}[Region count for parallel hyperplane families]
\label{prop:parallel-arrangement}
Let $r,m,\kappa\in\mathbb N$. An arrangement in $\R^r$ consisting of at most $m$ parallel families, with at most $\kappa$ hyperplanes in each family, has at most
\[
    A_r(m,\kappa):=\sum_{s=0}^{\min\{r,m\}}\binom ms \kappa^s
\]
full-dimensional regions. For $\kappa=1$, equality is attained by $m$ hyperplanes in general position. Define $A_0(m,\kappa):=1$.
\end{proposition}

\begin{remark}
The parameter $\kappa$ bounds the number of cuts within each parallel family, while $r$ limits the number of independent directions. The case $\kappa=1$ gives the usual hyperplane arrangement bound used in the ReLU comparison.
\end{remark}

\begin{example}
Two vertical lines and two horizontal lines divide $\R^2$ into $A_2(2,2)=1+\binom21 2+\binom22 2^2=9$ regions.
\end{example}

\subsection{Piecewise-Linear KANs}
\label{subsec:piecewise-linear-kans}

A KAN places univariate functions on its edges and sums their
values at each destination node
\cite{liu2025kan,schoots2025relating}. We study the case in which
these edge functions are CPWL.

\begin{definition}[Knot]
For a scalar CPWL function $\phi:\R\to\R$, a knot is a point $t\in\R$ such that $\phi$ is not affine on any open interval containing $t$. Its knot set is denoted by $\mathcal K(\phi)$. The connected components of $\R\setminus\mathcal K(\phi)$ are its maximal open affine intervals, whose number is $\abs{\mathcal K(\phi)}+1$ \cite{de1978practical,goujon2024number}.
\end{definition}

Fix an input dimension $d\in\mathbb N$, a number of layers
$T\in\mathbb N$, and an integer $k\geq2$ specifying the maximum number of affine segments per edge function. Let $K:=k-1$ denote the corresponding maximum number of knots per edge function. Let $\mathbf n=(n_0,\ldots,n_T)$ be the width vector, where $n_0=d$ and $n_\ell\in\mathbb N$ for $\ell\in[T]$. The network has $T-1$ hidden layers. Scalar-output networks have $n_T=1$.

\begin{definition}[CPWL KAN]
A CPWL KAN with architecture $\mathbf n$ uses scalar CPWL edge functions $\phi_{\ell,j,i}:\R\to\R$, each with at most $k$ affine segments, for $\ell\in[T]$, $i\in[n_{\ell-1}]$, and $j\in[n_\ell]$.
Let $z=(z_1,\ldots,z_{n_{\ell-1}})\in\R^{n_{\ell-1}}$ denote the input vector to layer $\ell$. The layer map
$\Phi_\ell:\R^{n_{\ell-1}}\to\R^{n_\ell}$ is defined by
\[
    [\Phi_\ell(z)]_j := \sum_{i=1}^{n_{\ell-1}}\phi_{\ell,j,i}(z_i),
    \quad j\in[n_\ell].
\]
Given a network input $x\in\R^d$, set $z^{(0)}:=x$ and define the representation after layer $\ell$ by $z^{(\ell)}:=\Phi_\ell(z^{(\ell-1)})$. The network output is $f(x):=z^{(T)}$. When $n_T=1$, we identify the output with the scalar $z_1^{(T)}$~\cite{liu2025kan,schoots2025relating}.
\end{definition}

Finite sums and compositions of CPWL maps are CPWL, so every network defined above is CPWL~\cite{goujon2024number}. Knot locations and affine coefficients may vary, provided that each edge function is continuous and has at most $k$ affine segments. Linear regions are counted over $\R^d$.

\begin{example}
For $\mathbf n=(2,1)$ and $k=2$, choosing both edge functions as $\phi(t)=\abs{t}$ gives $f(x_1,x_2)=\abs{x_1}+\abs{x_2}$. The coordinate lines $x_1=0$ and $x_2=0$ divide the input space into four linear regions.
\end{example}

We study CPWL KANs in which each edge may choose its knots independently. This class provides the setting for our comparison with Schoots et al.~\cite{schoots2025relating}. We also consider a structured subclass in which all edges leaving the same source node use a common candidate knot set. This restriction allows us to examine how knot sharing affects the maximum number of linear regions.

\begin{definition}[Edge-wise-knot CPWL KAN]
An edge-wise-knot CPWL KAN allows each edge function $\phi_{\ell,j,i}$ to choose its own knot set, subject to
$\abs{\mathcal K(\phi_{\ell,j,i})}\leq K$. The class of these networks with architecture $\mathbf n$ and at most $k$ affine segments per edge is denoted by $\mathcal F_{\mathrm{ind}}(\mathbf n,k)$.
\end{definition}

\begin{definition}[Source-shared-knot CPWL KAN]
A source-shared-knot CPWL KAN assigns a candidate knot set
$\mathcal S_{\ell,i}\subset\R$, with
$\abs{\mathcal S_{\ell,i}}\leq K$, to every source
$i\in[n_{\ell-1}]$ in every layer $\ell\in[T]$.
Every outgoing edge satisfies
$\mathcal K(\phi_{\ell,j,i})\subseteq\mathcal S_{\ell,i}$
for $j\in[n_\ell]$, while its affine coefficients remain
edge-specific. This class is denoted by
$\mathcal F_{\mathrm{sh}}(\mathbf n,k)$, and
$\mathcal F_{\mathrm{sh}}(\mathbf n,k)
\subseteq\mathcal F_{\mathrm{ind}}(\mathbf n,k)$.
\end{definition}

\begin{example}
Consider a layer with one input and two outputs, with at most $K=2$ knots per edge. In the edge-wise-knot model, the two edges may use the knot sets $\{0,1\}$ and $\{2,3\}$. These four distinct knots divide the input line into five linear regions. In the source-shared-knot model, choose the common candidate set $\mathcal S_1=\{0,1\}$. The first edge may use both knots and
the second only $1$, giving three linear regions. Thus, sharing limits the available knot locations while allowing each edge to use different subsets and affine coefficients.
\end{example}

\section{Maximum Number of Linear Regions of CPWL KANs}
\label{sec:maximum-regions}

In this section, we establish matching upper and lower bounds on the maximum number of linear regions of CPWL KANs. We first count the regions created by one layer, then derive upper bounds for deep networks and construct matching lower bounds for arbitrary architectures. Together, these results determine the tight growth order in the number of affine segments per edge, including the effects of dimensional bottlenecks.

\subsection{Regions Created by One KAN Layer}

Consider a KAN layer $\Phi:\R^p\to\R^q$ given by $[\Phi(z)]_j=\sum_{i=1}^{p}\phi_{j,i}(z_i),\quad j\in[q]$. For each source coordinate $i\in[p]$, define the union of the knots on its outgoing edges by
\[
    \mathcal{K}^{\mathrm{out}}_i:=\bigcup_{j=1}^{q}\mathcal{K}(\phi_{j,i}),\quad M_i:=\abs{\mathcal{K}^{\mathrm{out}}_i}.
\]

\begin{proposition}[Exact region count of one KAN layer]
\label{prop:one-layer-kan}
The polyhedral partition induced by $\Phi$ consists of axis-aligned open boxes and contains exactly $\prod_{i=1}^{p}(M_i+1)$ full-dimensional cells. Consequently, the maximum numbers of cells
in the edge-wise-knot and source-shared-knot models are $\max_{\Phi\in\mathcal{F}_{\mathrm{ind}}((p,q),k)}\mathcal{R}(\Phi) = \bigl(q(k-1)+1\bigr)^p$ and $\max_{\Phi\in\mathcal{F}_{\mathrm{sh}}((p,q),k)}\mathcal{R}(\Phi)=k^p$, respectively.
\end{proposition}

\begin{remark}
Every knot associated with source coordinate $z_i$ produces a cut
of the form $z_i=\tau$. Thus, $M_i$ knots divide the $i$th axis into
$M_i+1$ intervals, and taking one interval from every coordinate
produces an axis-aligned box. In the edge-wise model, the $q$
outgoing edges may contribute $q(k-1)$ distinct knots, whereas the
source-shared model has at most $k-1$ distinct knots per source.
\end{remark}

\subsection{Upper Bounds on the Maximum Number of Linear Regions}

Let $F_\ell:=\Phi_\ell\circ\cdots\circ\Phi_1$ and $F_0(x):=x$
denote the partial network after layer $\ell\in[T]$. For each
$\ell\in[T]$, define the effective dimension before layer $\ell$ by $\rho_{\ell-1}:=\min_{0\leq s\leq\ell-1}n_s$.

\begin{lemma}[Subdivision produced by one layer]
\label{lem:layerwise-subdivision}
Let $P\subset\R^d$ be a full-dimensional polyhedral cell on which $F_{\ell-1}(x)=W_Px+c_P$, where $W_P\in\R^{n_{\ell-1}\times d}$ and
$c_P\in\R^{n_{\ell-1}}$. Let $r_P:=\operatorname{rank}(W_P)$. Then layer $\ell$ divides $P$ into at most $A_{r_P}\bigl(n_{\ell-1},n_\ell(k-1)\bigr)$ full-dimensional cells in the edge-wise-knot model and at most $A_{r_P}\bigl(n_{\ell-1},k-1\bigr)$ full-dimensional cells in the source-shared-knot model. Moreover, $r_P\leq\rho_{\ell-1}$. \end{lemma}

\begin{remark}
Inside $P$, all preceding layers act as one affine map. A knot
$\tau$ associated with source coordinate $i$ therefore pulls back
to the hyperplane $\left\{x\in P:(W_P)_{i,:}x+(c_P)_i=\tau\right\}$.
Knots associated with the same source coordinate have the same
normal vector and hence form a parallel family. The rank $r_P$
gives the number of independent directions in which these
hyperplanes can divide $P$.
\end{remark}

For $\diamond\in\{\mathrm{ind},\mathrm{sh}\}$, let $\mathcal{R}^{\star}_{\diamond}(\mathbf n,k):=\max_{f\in\mathcal{F}_{\diamond}(\mathbf n,k)}\mathcal{R}(f)$ denote the maximum number of linear regions among networks with architecture $\mathbf n=(n_0,\ldots,n_T)$ and at most $k$ affine segments per edge.

Applying Lemma~\ref{lem:layerwise-subdivision} successively to all
cells produced by the preceding layers gives the following
network-wide upper bounds.

\begin{theorem}[Upper bounds on the maximum number of linear regions]
\label{thm:global-upper}
Let $n_0=d$ and $n_T=1$. Then
\[
\begin{aligned}
    \mathcal R^{\star}_{\mathrm{ind}}(\mathbf n,k)
    &\leq
    \prod_{\ell=1}^{T}
    \left[
        \sum_{s=0}^{\rho_{\ell-1}}
        \binom{n_{\ell-1}}{s}
        \bigl(n_\ell(k-1)\bigr)^s
    \right],\\
    \mathcal R^{\star}_{\mathrm{sh}}(\mathbf n,k)
    &\leq
    \prod_{\ell=1}^{T}
    \left[
        \sum_{s=0}^{\rho_{\ell-1}}
        \binom{n_{\ell-1}}{s}
        (k-1)^s
    \right].
\end{aligned}
\]
These bounds imply
\[
\begin{aligned}
    \mathcal R^{\star}_{\mathrm{ind}}(\mathbf n,k)
    &\leq
    \left\{
        \prod_{\ell=1}^{T}
        \left[
            \sum_{s=0}^{\rho_{\ell-1}}
            \binom{n_{\ell-1}}{s}n_\ell^s
        \right]
    \right\}
    k^{\sum_{\ell=1}^{T}\rho_{\ell-1}},\\
    \mathcal R^{\star}_{\mathrm{sh}}(\mathbf n,k)
    &\leq
    \left\{
        \prod_{\ell=1}^{T}
        \left[
            \sum_{s=0}^{\rho_{\ell-1}}
            \binom{n_{\ell-1}}{s}
        \right]
    \right\}
    k^{\sum_{\ell=1}^{T}\rho_{\ell-1}}.
\end{aligned}
\]
Consequently, for every fixed architecture $\mathbf n$,
as $k\to\infty$,
\[
\begin{aligned}
    \mathcal R^{\star}_{\mathrm{ind}}(\mathbf n,k)
    &=
    O_{\mathbf n}\!\left(
        k^{\sum_{\ell=1}^{T}\rho_{\ell-1}}
    \right),\\
    \mathcal R^{\star}_{\mathrm{sh}}(\mathbf n,k)
    &=
    O_{\mathbf n}\!\left(
        k^{\sum_{\ell=1}^{T}\rho_{\ell-1}}
    \right).
\end{aligned}
\]
\end{theorem}

\begin{remark}
Before layer $\ell$, the representation has at most
$\rho_{\ell-1}$ independent directions on each cell.
This layer contributes at most a factor of order
$k^{\rho_{\ell-1}}$, giving the total exponent
$\sum_{\ell=1}^{T}\rho_{\ell-1}$.
The displayed upper bounds hold for every integer $k\geq2$.
For the asymptotic statements, the architecture $\mathbf n$
is fixed and $k$ increases. The width-dependent multiplicative
factors are therefore constant, while the displayed power of
$k$ determines the growth rate. Layer widths affect both
these factors and the exponent through the effective
dimensions $\rho_{\ell-1}$.
\end{remark}

\subsection{Matching Lower Bounds on the Maximum Number of Linear Regions}

Theorem~\ref{thm:global-upper} bounds the growth in $k$
through the effective dimensions of successive layers.
We next construct a source-shared-knot CPWL KAN that attains
the same exponent for every architecture, including those
with hidden widths smaller than $d$.
Since the source-shared-knot class is contained in the
edge-wise-knot class, this construction gives a lower
bound for both models.

\begin{theorem}[Lower bounds on the maximum number of linear regions]
\label{thm:global-lower}
Let $n_0=d$ and $n_T=1$. Then
\[
    \mathcal R^{\star}_{\mathrm{ind}}(\mathbf n,k)
    \geq
    \mathcal R^{\star}_{\mathrm{sh}}(\mathbf n,k)
    \geq
    k^{\sum_{\ell=1}^{T}\rho_{\ell-1}}.
\]
\end{theorem}

\begin{remark}
Each layer applies a $k$-piece folding function to every
active coordinate and averages groups of the folded coordinates.
Every existing cell maps onto the same cube of the next
effective dimension, allowing each layer to create
$k^{\rho_{\ell-1}}$ child cells.
The chosen branch slopes give different final gradients
on all counted cells, so these cells belong to different
linear regions.
\end{remark}

The upper and lower bounds have the same power of $k$.
Combining them determines the asymptotic order for every
fixed architecture.

\begin{corollary}[Tight asymptotic bounds]
\label{cor:tight-kan-bounds}
For every fixed architecture $\mathbf n$ with $n_0=d$
and $n_T=1$, as $k\to\infty$,
\[
\begin{aligned}
    \mathcal R^{\star}_{\mathrm{sh}}(\mathbf n,k)
    &=
    \Theta_{\mathbf n}\!\left(
        k^{\sum_{\ell=1}^{T}\rho_{\ell-1}}
    \right),\\
    \mathcal R^{\star}_{\mathrm{ind}}(\mathbf n,k)
    &=
    \Theta_{\mathbf n}\!\left(
        k^{\sum_{\ell=1}^{T}\rho_{\ell-1}}
    \right).
\end{aligned}
\]
The implicit constants depend only on $\mathbf n$
and are independent of $k$.
\end{corollary}

\begin{remark}
Both knot parameterizations have the same asymptotic exponent
for every fixed architecture. If every hidden width is at
least $d$, then $\rho_{\ell-1}=d$ for all $\ell\in[T]$,
and both maximum region counts have order
$\Theta_{\mathbf n}(k^{dT})$.
For architectures with narrower hidden layers, the exponent
is determined by the reduced effective dimensions.
\end{remark}

\section{Comparison with Existing KAN Bounds and ReLU Networks}
\label{sec:comparison}

In this section, we use the bounds established in
Section~\ref{sec:maximum-regions} to quantify the improvement
over the KAN upper bound of
Schoots et al.~\cite{schoots2025relating}. We then describe
exact conversions between CPWL KANs and ReLU networks and
compare their maximum numbers of linear regions under a common
parameter budget and the same fixed number of nonlinear layers.

Schoots et al.~\cite{schoots2025relating} obtained the
edgewise product bound $\mathcal R(f) \leq k^{\sum_{\ell=1}^{T}n_{\ell-1}n_\ell}$. For scalar output, $n_0=d$ and $n_T=1$. Thus, for $T\geq2$, their bound becomes $\mathcal R(f) \leq k^{dn_1+\sum_{\ell=1}^{T-2}n_\ell n_{\ell+1} +n_{T-1}}$.
If every hidden layer has width $n$, this expression reduces to $\mathcal R(f)\leq k^{dn+(T-2)n^2+n}$.

The following corollary compares this result with our finite
upper bounds and the tight asymptotic order established in
Corollary~\ref{cor:tight-kan-bounds}.

\begin{corollary}[Comparison with the KAN bound of \citep{schoots2025relating}]
\label{cor:aistats-comparison}
For every architecture $\mathbf n$ and every $k\geq2$,
\[
\begin{aligned}
    &\prod_{\ell=1}^{T}
    \left[
        \sum_{s=0}^{\rho_{\ell-1}}
        \binom{n_{\ell-1}}s(k-1)^s
    \right] \leq
    \prod_{\ell=1}^{T}
    \left[
        \sum_{s=0}^{\rho_{\ell-1}}
        \binom{n_{\ell-1}}s
        \bigl(n_\ell(k-1)\bigr)^s
    \right] \leq
    k^{\sum_{\ell=1}^{T}n_{\ell-1}n_\ell}.
\end{aligned}
\]
For every fixed architecture $\mathbf n$ and each
$\diamond\in\{\mathrm{ind},\mathrm{sh}\}$,
\[
    \frac{
        \mathcal R^\star_{\diamond}(\mathbf n,k)
    }{
        k^{\sum_{\ell=1}^{T}n_{\ell-1}n_\ell}
    }
    =
    \Theta_{\mathbf n}\!\left(
        k^{
            \sum_{\ell=1}^{T}\rho_{\ell-1}
            -
            \sum_{\ell=1}^{T}n_{\ell-1}n_\ell
        }
    \right)
    \quad\text{as }k\to\infty.
\]
\end{corollary}

\begin{remark}
The bound of Schoots et al.~\cite{schoots2025relating}
counts segment choices separately for every edge.
Our bound accounts for the parallel cuts associated with each
source coordinate and the effective dimension before each layer.
This reduces the exponent from
$\sum_{\ell=1}^{T}n_{\ell-1}n_\ell$
to the tight exponent
$\sum_{\ell=1}^{T}\rho_{\ell-1}$. The difference between these exponents satisfies
\[
\begin{aligned}
    &\sum_{\ell=1}^{T}n_{\ell-1}n_\ell
    -
    \sum_{\ell=1}^{T}\rho_{\ell-1}=
    \sum_{\ell=1}^{T}
    \left[
        n_{\ell-1}-\rho_{\ell-1}
        +
        n_{\ell-1}(n_\ell-1)
    \right]
    \geq0.
\end{aligned}
\]
For $T\geq2$, equality holds exactly when every hidden
layer has width one. Hence, whenever at least one hidden
layer has width greater than one, our upper bound is
asymptotically strictly tighter and the ratio in the
corollary converges to zero. If every hidden width equals
one, both bounds are exact and give $k^{d+T-1}$.
For $T=1$, both bounds are exact and give $k^d$.

For a common hidden width $n$ and $T\geq2$, $\sum_{\ell=1}^{T}\rho_{\ell-1}=d+(T-1)\min\{d,n\}$. Consequently,
\[
    \frac{
        \mathcal R^\star_{\diamond}(\mathbf n,k)
    }{
        k^{dn+(T-2)n^2+n}
    }
    =
    \Theta_{d,n,T}\!\left(
        k^{
            d+(T-1)\min\{d,n\}
            -dn-(T-2)n^2-n
        }
    \right).
\]
When $n\geq d$, the tight exponent
$d+(T-1)\min\{d,n\}$ simplifies to $dT$.
\end{remark}

\begin{example}
Consider the architecture $\mathbf n=(2,2,1)$, with input dimension $d=2$ and $T=2$ layers. Our bounds give
\[
\begin{aligned}
    \mathcal R^\star_{\mathrm{sh}}(\mathbf n,k)
    &=k^4,\\
    k^4
    \leq \mathcal R^\star_{\mathrm{ind}}(\mathbf n,k)
    &\leq k^2(2k-1)^2.
\end{aligned}
\]
The upper bound of~\citep{schoots2025relating} is $k^{2\cdot2+2\cdot1}=k^6$. Its ratio to our edge-wise-knot  upper bound is
\[
    \frac{k^6}{k^2(2k-1)^2}=\frac{k^4}{(2k-1)^2}=\Theta(k^2).
\]
For $k=10$, our lower and upper bounds for the edge-wise-knot model are $10\,000$ and $36\,100$, respectively. The upper bound of \citep{schoots2025relating}. is $1\,000\,000$, approximately $27.7$ times our upper bound. Thus, our bounds determine the attainable growth order $k^4$, and the improvement over the existing upper bound grows quadratically with $k$.
\end{example}

The preceding comparison concerns bounds for CPWL KANs.
We now compare CPWL KANs with ReLU networks under the same
total parameter budget. We first recall their functional
relationship.

\begin{proposition}[Exact conversion between KAN and ReLU networks]
\label{prop:kan-relu-equivalence}
Let $p,q\in\mathbb N$. A CPWL KAN layer from $\R^p$ to
$\R^q$ with at most $K=k-1$ knots per edge can be represented
exactly by a ReLU network with one hidden layer and an affine
output layer. The hidden width is at most $pqK+2p$ for
edge-wise knots and at most $pK+2p$ for source-shared knots.
Conversely, a ReLU layer from $\R^p$ to $\R^q$ can be
represented exactly by two source-shared-knot KAN layers
with widths $(p,q,q)$ and at most two affine segments
per edge ($k=2$).
\end{proposition}

\begin{remark}
Each CPWL edge function is an affine function plus a sum of
shifted ReLU functions. This representation gives the conversion,
but changes the network width or depth. Parameter efficiency
therefore requires a separate comparison.
\end{remark}

We now compare the two models under a common total parameter
budget $\Lambda\in\mathbb N$. We use the edge-wise-knot KAN
class and align the number of nonlinear layers: a KAN has
$T$ KAN layers, while a ReLU network has $T$ ReLU hidden
layers followed by an affine scalar output layer. Both models
have input dimension $d$ and scalar output.

In the shifted-ReLU representation, each KAN edge has two
affine coefficients, $K$ slope-change coefficients, and $K$
trainable candidate knot locations. Its parameter count is
therefore $2+2K=2k$. For the KAN architecture $\mathbf n$,
the total parameter count is
\[
    \Lambda_{\mathrm{KAN}}(\mathbf n,k):=2k\sum_{\ell=1}^{T}n_{\ell-1}n_\ell.
\]
For the ReLU network, let
$\mathbf m=(m_0,\ldots,m_{T+1})$ be its width vector, where
$m_0=d$, $m_{T+1}=1$, and $m_1,\ldots,m_T\in\mathbb N$.
Counting all weights and biases gives
\[
    \Lambda_{\mathrm{ReLU}}(\mathbf m)
    :=
    \sum_{\ell=1}^{T+1}(m_{\ell-1}+1)m_\ell.
\]
Both counts include every entry of the fully connected
parameterization, including entries whose values are zero.

For sufficiently large $\Lambda$, let
$\mathcal R^\star_{\mathrm{KAN}}(d,T,\Lambda)$ and
$\mathcal R^\star_{\mathrm{ReLU}}(d,T,\Lambda)$ denote
the maximum numbers of linear regions attainable by the
respective models using at most $\Lambda$ parameters.
Widths and $k$ may vary, provided that the total number
of parameters does not exceed $\Lambda$.

\begin{theorem}[Comparison under equal parameter budgets]
\label{thm:parameter-budget-comparison}
For fixed $d,T\in\mathbb N$ and $\Lambda\to\infty$,
\[
\begin{aligned}
    \mathcal R^\star_{\mathrm{KAN}}(d,T,\Lambda)
    &=
    \Theta_{d,T}\!\left(\Lambda^{dT}\right),\\
    \mathcal R^\star_{\mathrm{ReLU}}(d,T,\Lambda)
    &=
    \Theta_{d,T}\!\left(
        \Lambda^{d\lceil T/2\rceil}
    \right).
\end{aligned}
\]
Consequently,
\[
    \frac{
        \mathcal R^\star_{\mathrm{KAN}}(d,T,\Lambda)
    }{
        \mathcal R^\star_{\mathrm{ReLU}}(d,T,\Lambda)
    }
    =
    \Theta_{d,T}\!\left(
        \Lambda^{d\lfloor T/2\rfloor}
    \right).
\]
The implicit constants depend only on $d$ and $T$.
\end{theorem}

\begin{remark}
For $T\geq2$, the ratio tends to infinity. Thus, under the
stated parameter count and fixed nonlinear depth, CPWL KANs
attain asymptotically more linear regions. A KAN can keep its
widths fixed and spend additional parameters on knots in every
layer. In a fully connected ReLU network, increasing adjacent
widths also increases the number of connections between them.
The KAN advantage here allows $k$ to grow with $\Lambda$.
\end{remark}

\section{Conclusion}
\label{sec:conclusion}

In this paper, we established tight asymptotic bounds for the maximum number of linear regions of CPWL KANs. For every fixed architecture, both knot parameterizations satisfy
\[
    \mathcal R^\star_\diamond(\mathbf n,k)
    =
    \Theta_{\mathbf n}\!\left(
        k^{\sum_{\ell=1}^{T}\rho_{\ell-1}}
    \right),
    \quad
    \diamond\in\{\mathrm{ind},\mathrm{sh}\}.
\]
This result quantifies the effect of dimensional bottlenecks and sharpens the edgewise product bound of Schoots et al.~\cite{schoots2025relating}. Under the dense parameter counts and fixed nonlinear depth considered here, allowing $k$ to vary also gives CPWL KANs a larger asymptotic maximum region count than ReLU networks for $T\geq2$. These results connect local knot placement and network architecture to a precise law for the growth of geometric expressive capacity.

\bibliography{references}

@article{goujon2024number,
  title={On the number of regions of piecewise linear neural networks},
  author={Goujon, Alexis and Etemadi, Arian and Unser, Michael},
  journal={Journal of Computational and Applied Mathematics},
  volume={441},
  pages={115667},
  year={2024},
  publisher={Elsevier}
}

@book{de1978practical,
  title={A practical guide to splines},
  author={De Boor, Carl and De Boor, Carl},
  volume={27},
  year={1978},
  publisher={springer, New York}
}

@book{zaslavsky1975facing,
  title={Facing up to arrangements: Face-count formulas for partitions of space by hyperplanes: Face-count formulas for partitions of space by hyperplanes},
  author={Zaslavsky, Thomas},
  volume={154},
  year={1975},
  publisher={American Mathematical Soc.}
}

@article{montufar2014number,
  title={On the number of linear regions of deep neural networks},
  author={Mont{\'u}far, Guido and Pascanu, Razvan and Cho, Kyunghyun and Bengio, Yoshua},
  journal={Advances in neural information processing systems},
  volume={27},
  year={2014}
}

@inproceedings{serra2018bounding,
  title={Bounding and counting linear regions of deep neural networks},
  author={Serra, Thiago and Tjandraatmadja, Christian and Ramalingam, Srikumar},
  booktitle={International conference on machine learning},
  pages={4558--4566},
  year={2018},
  organization={PMLR}
}

@inproceedings{liu2025kan,
  title={KAN: Kolmogorov--arnold networks},
  author={Liu, Ziming and Wang, Yixuan and Vaidya, Sachin and Ruehle, Fabian and Halverson, James and Soljacic, Marin and Hou, Thomas and Tegmark, Max},
  booktitle={International conference on learning representations},
  volume={2025},
  pages={70367--70413},
  year={2025}
}

@inproceedings{schoots2025relating,
  author    = {Schoots, Nandi and Villani, Mattia Jacopo and
               uit de Bos, Niels},
  title     = {Relating Piecewise Linear Kolmogorov Arnold Networks
               to {ReLU} Networks},
  booktitle = {Proceedings of the 28th International Conference
               on Artificial Intelligence and Statistics},
  series    = {Proceedings of Machine Learning Research},
  volume    = {258},
  pages     = {199--207},
  year      = {2025},
  publisher = {PMLR}
}

@article{su2026expressivity,
  title={Expressivity of Transformers: A Tropical Geometry Perspective},
  author={Su, Ye and Liu, Yong},
  journal={arXiv preprint arXiv:2604.14727},
  year={2026}
}

@article{su2026sparsity,
  title={Sparsity is combinatorial depth: Quantifying moe expressivity via tropical geometry},
  author={Su, Ye and Tang, Huayi and Gong, Zixuan and Liu, Yong},
  journal={arXiv preprint arXiv:2602.03204},
  year={2026}
}

@article{xiong2024number,
  title={On the number of linear regions of convolutional neural networks with piecewise linear activations},
  author={Xiong, Huan and Huang, Lei and Zang, Wenston JT and Zhen, Xiantong and Xie, Guo-Sen and Gu, Bin and Song, Le},
  journal={IEEE Transactions on Pattern Analysis and Machine Intelligence},
  volume={46},
  number={7},
  pages={5131--5148},
  year={2024},
  publisher={IEEE}
}

@inproceedings{xiong2020number,
  title={On the number of linear regions of convolutional neural networks},
  author={Xiong, Huan and Huang, Lei and Yu, Mengyang and Liu, Li and Zhu, Fan and Shao, Ling},
  booktitle={International Conference on Machine Learning},
  pages={10514--10523},
  year={2020},
  organization={PMLR}
}

@article{hu2020analysis,
  title={Analysis on the number of linear regions of piecewise linear neural networks},
  author={Hu, Qiang and Zhang, Hao and Gao, Feifei and Xing, Chengwen and An, Jianping},
  journal={IEEE transactions on neural networks and learning systems},
  volume={33},
  number={2},
  pages={644--653},
  year={2020},
  publisher={IEEE}
}

@inproceedings{hanin2019complexity,
  title={Complexity of linear regions in deep networks},
  author={Hanin, Boris and Rolnick, David},
  booktitle={International Conference on Machine Learning},
  pages={2596--2604},
  year={2019},
  organization={PMLR}
}

@article{somvanshi2025survey,
  title={A survey on kolmogorov-arnold network},
  author={Somvanshi, Shriyank and Javed, Syed Aaqib and Islam, Md Monzurul and Pandit, Diwas and Das, Subasish},
  journal={ACM Computing Surveys},
  volume={58},
  number={2},
  pages={1--35},
  year={2025},
  publisher={ACM, New York, NY}
}

@article{dutta2026first,
  title={The first two months of kolmogorov-arnold networks (kans): a survey of the state-of-the-art},
  author={Dutta, Anurag and Maheswari, B and Punitha, N and Raj, A Stephan Antony and Banu, S Sharmila and Balamurugan, M},
  journal={Archives of Computational Methods in Engineering},
  volume={33},
  number={1},
  pages={1017--1028},
  year={2026},
  publisher={Springer}
}

@article{montufar2022sharp,
  author  = {Mont{\'u}far, Guido and Ren, Yue and Zhang, Leon},
  title   = {Sharp Bounds for the Number of Regions of
             Maxout Networks and Vertices of Minkowski Sums},
  journal = {SIAM Journal on Applied Algebra and Geometry},
  volume  = {6},
  number  = {4},
  pages   = {618--649},
  year    = {2022},
  doi     = {10.1137/21M1413699}
}

@article{chen2023lower,
  author  = {Chen, Hao and Wang, Yu Guang and Xiong, Huan},
  title   = {Lower and Upper Bounds for Numbers of Linear
             Regions of Graph Convolutional Networks},
  journal = {Neural Networks},
  volume  = {168},
  pages   = {394--404},
  year    = {2023},
  doi     = {10.1016/j.neunet.2023.09.025}
}

@inproceedings{pham2024what,
  author    = {Pham, Tuan Anh and Garg, Vikas},
  title     = {What Do Graph Neural Networks Learn?
               Insights from Tropical Geometry},
  booktitle = {Advances in Neural Information Processing Systems},
  volume    = {37},
  year      = {2024},
  url       = {https://proceedings.neurips.cc/paper_files/paper/2024/hash/14d2d59c5e2302bfa6ce6cae59156e33-Abstract-Conference.html}
}

@inproceedings{bakaev2026better,
  author    = {Bakaev, Egor and Brunck, Florestan
               and Hertrich, Christoph and Stade, Jack
               and Yehudayoff, Amir},
  title     = {Better Neural Network Expressivity:
               Subdividing the Simplex},
  booktitle = {Proceedings of the 58th Annual ACM Symposium
               on Theory of Computing},
  pages     = {500--507},
  publisher = {ACM},
  year      = {2026},
  doi       = {10.1145/3798129.3800768}
}

\appendix

\section{Proofs}
\label{app:proofs}

Throughout this appendix, let $K:=k-1$ and $\sigma(t):=\max\{0,t\}$. We identify full-dimensional polyhedral cells with their open interiors and use the conventions that empty sums equal $0$ and empty products equal $1$. Table~\ref{tab:notation} summarizes the notation used in the main text and proofs, with context-specific meanings indicated where a symbol is reused.

\begingroup
\small
\setlength{\tabcolsep}{5pt}
\renewcommand{\arraystretch}{1.12}
\setlength{\LTleft}{0pt}
\setlength{\LTright}{0pt}
\setlength{\LTcapwidth}{\textwidth}

\begin{longtable}{@{}p{0.31\textwidth}p{0.65\textwidth}@{}}
\caption{Notation used in the main text and proofs.}
\label{tab:notation}\\
\toprule
Symbol & Meaning\\
\midrule
\endfirsthead

\multicolumn{2}{@{}l@{}}{
    \textit{Table~\thetable\ continued}
}\\
\toprule
Symbol & Meaning\\
\midrule
\endhead

\midrule
\multicolumn{2}{r@{}}{
    \textit{Continued on the next page}
}\\
\endfoot

\bottomrule
\endlastfoot

\multicolumn{2}{@{}l@{}}{
    \textbf{Basic notation and polyhedral geometry}
}\\
\addlinespace[0.3em]

$\R$, $\mathbb N$
&
The real numbers and the positive integers, respectively.
\\

$[m]$
&
The index set $\{1,\ldots,m\}$ for $m\in\mathbb N$.
\\

$\abs{S}$, $\abs{t}$
&
The cardinality of a finite set $S$ and the absolute value
of a real number $t$, respectively.
\\

$x$, $z$, $t$
&
An input vector, an intermediate layer input, and a scalar
argument, respectively.
\\

$p$, $q$
&
The input and output dimensions of the map under
consideration, respectively. For a network layer,
these are its input and output widths.
\\

$A$, $b$
&
The matrix and offset vector of an affine map
$x\mapsto Ax+b$.
\\

$C$, $v$, $s$
&
The constraint matrix, constraint vector, and number of
inequalities in a polyhedron $P=\{x:Cx\leq v\}$.
\\

$\mathcal P$, $P_a$, $M$
&
A finite polyhedral partition, its $a$-th cell, and its
number of cells, respectively.
\\

$\operatorname{int}(P)$
&
The interior of $P$ in its ambient Euclidean space.
\\

$g$, $A_a$, $b_a$
&
A CPWL map and the coefficients of its affine formula
$g(x)=A_ax+b_a$ on $P_a$.
\\

$U$, $\mathcal R(g)$
&
A linear region of $g$ and the number of its linear regions.
A linear region is an inclusion-maximal nonempty connected
open set on which $g$ agrees with one affine map.
\\

$H(u,\beta)$
&
The hyperplane $\{x:u^{\mathsf T}x=\beta\}$, where
$u$ is a nonzero normal vector and $\beta$ is its offset.
\\

$\mathcal H$
&
A finite hyperplane arrangement.
\\

$r$, $m$, $\kappa$
&
In $A_r(m,\kappa)$, the ambient dimension, maximum number
of parallel families, and maximum number of hyperplanes
per family, respectively.
\\

$A_r(m,\kappa)$
&
The arrangement bound
$\sum_{s=0}^{\min\{r,m\}}\binom ms \kappa^s$,
with $A_0(m,\kappa)=A_r(0,\kappa)=1$.
\\

$\binom ms$
&
The binomial coefficient, taken to be zero when
$s<0$ or $s>m$.
\\

$\operatorname{rank}(A)$,
$\operatorname{row}(A)$
&
The rank and row space of a matrix $A$, respectively.
\\

$\dim(V)$, $I_d$
&
The dimension of a vector or affine space $V$ and the
$d\times d$ identity matrix, respectively.
\\

$(\cdot)^{\mathsf T}$, $\circ$
&
Matrix transpose and composition of maps, respectively.
\\

$g|_P$, $g^{-1}(S)$
&
The restriction of $g$ to $P$ and the preimage of a set
$S$ under $g$, respectively.
\\

$\bigsqcup$
&
Disjoint union.
\\

$\nabla f$, $\partial_{x_h}f$
&
The gradient of $f$ and its partial derivative with respect
to input coordinate $x_h$, respectively.
\\

$\partial_{z_i}^{-}$,
$\partial_{z_i}^{+}$
&
The left and right derivatives along coordinate $z_i$,
respectively.
\\

$\lfloor t\rfloor$,
$\lceil t\rceil$
&
The greatest integer not exceeding $t$ and the least
integer not smaller than $t$, respectively.
\\

$O_{\mathbf n}(\cdot)$,
$\Theta_{\mathbf n}(\cdot)$
&
An asymptotic upper bound and matching upper and lower
asymptotic orders. The constants may depend on the fixed
architecture $\mathbf n$ but are independent of $k$.
Other subscripts specify the allowed parameter dependence.
\\

\addlinespace[0.5em]
\multicolumn{2}{@{}l@{}}{
    \textbf{CPWL KAN architecture}
}\\
\addlinespace[0.3em]

$d$, $T$
&
The input dimension and number of KAN layers, respectively.
A KAN has $T-1$ hidden layers. In the comparison with ReLU networks, $T$ also denotes the number of ReLU hidden layers.
\\

$k$, $K$
&
The maximum numbers of affine segments and actual knots
per edge function, respectively, with $K=k-1$.
\\

$\mathbf n$, $n_\ell$
&
The KAN width vector $\mathbf n=(n_0,\ldots,n_T)$
and its layer widths, with $n_0=d$ and $n_T=1$
for scalar output.
\\

$\ell$, $i$, $j$
&
The layer, source-node, and destination-node indices
in $\phi_{\ell,j,i}$, respectively.
\\

$\phi$, $\mathcal K(\phi)$
&
A scalar CPWL function and its set of actual knots.
\\

$\phi_{\ell,j,i}$
&
The edge function from source node $i$ in layer
$\ell-1$ to destination node $j$ in layer $\ell$.
\\

$\Phi_\ell$
&
The layer map defined by
$[\Phi_\ell(z)]_j
=\sum_{i=1}^{n_{\ell-1}}\phi_{\ell,j,i}(z_i)$.
\\

$z^{(\ell)}$, $z_i^{(\ell)}$
&
The representation after layer $\ell$ and its $i$-th
coordinate, with $z^{(0)}=x$.
\\

$f$
&
The network output. For scalar output,
$f(x)=z_1^{(T)}(x)$.
\\

$\mathcal S_{\ell,i}$
&
The common candidate knot set for all edges leaving
source node $i$ in layer $\ell$, with
$\abs{\mathcal S_{\ell,i}}\leq K$.
\\

$\mathcal S_i$
&
The corresponding candidate knot set for a single layer.
\\

$\mathcal F_{\mathrm{ind}}(\mathbf n,k)$
&
The class of edge-wise-knot CPWL KANs with architecture
$\mathbf n$ and at most $k$ affine segments per edge.
\\

$\mathcal F_{\mathrm{sh}}(\mathbf n,k)$
&
The source-shared-knot subclass with the same architecture
and at most $k$ affine segments per edge.
\\

$\diamond$
&
A model index in $\{\mathrm{ind},\mathrm{sh}\}$.
\\

$\mathcal R^\star_{\diamond}(\mathbf n,k)$
&
The maximum number of linear regions over
$\mathcal F_{\diamond}(\mathbf n,k)$.
\\

$F_\ell$, $F_0$
&
The partial network
$F_\ell=\Phi_\ell\circ\cdots\circ\Phi_1$
and the identity map $F_0(x)=x$.
\\

$\rho_{\ell-1}$, $\rho_T$
&
The effective-dimension bound before layer $\ell$,
$\rho_{\ell-1}=\min_{0\leq s\leq\ell-1}n_s$.
The lower-bound construction extends this notation
by setting $\rho_T=1$.
\\

$\displaystyle
\sum_{\ell=1}^{T}\rho_{\ell-1}$
&
The tight exponent of $k$ in the maximum region count
for a fixed architecture. It equals $dT$ when every
hidden width is at least $d$.
\\

\addlinespace[0.5em]
\multicolumn{2}{@{}l@{}}{
    \textbf{Single-layer counts and layerwise subdivision}
}\\
\addlinespace[0.3em]

$\Phi$, $\phi_{j,i}$
&
A single KAN layer from $\R^p$ to $\R^q$ and its
edge functions.
\\

$\mathcal K_i^{\mathrm{out}}$, $M_i$
&
The union
$\bigcup_{j=1}^{q}\mathcal K(\phi_{j,i})$
of knots leaving source coordinate $i$ and its cardinality.
\\

$\tau_{i,r}$
&
The $r$-th ordered knot in
$\mathcal K_i^{\mathrm{out}}$, with
$\tau_{i,0}=-\infty$ and
$\tau_{i,M_i+1}=+\infty$.
\\

$I_{i,r}$
&
The interval $(\tau_{i,r},\tau_{i,r+1})$ between
consecutive knots associated with source coordinate $i$.
\\

$\bm r$, $B_{\bm r}$
&
An interval-index tuple $\bm r=(r_1,\ldots,r_p)$
and the box
$B_{\bm r}=I_{1,r_1}\times\cdots\times I_{p,r_p}$.
\\

$a_{j,i,r_i}$,
$b_{j,i,r_i}$
&
The slope and intercept of $\phi_{j,i}$ on $I_{i,r_i}$.
\\

$P$, $W_P$, $c_P$
&
A full-dimensional open convex cell and the coefficients
of $F_{\ell-1}(x)=W_Px+c_P$ on $P$.
\\

$r_P$
&
The local rank $\operatorname{rank}(W_P)$.
\\

$w_{P,i}^{\mathsf T}$,
$(c_P)_i$
&
The $i$-th row of $W_P$ and the $i$-th coordinate
of $c_P$, respectively.
\\

$H_{P,i,\tau}$
&
The knot preimage
$\{x\in P:w_{P,i}^{\mathsf T}x+(c_P)_i=\tau\}$.
It creates a cut only when it is a proper hyperplane
section of $P$.
\\

$B_s$
&
The affine slope matrix of the selected pieces in
preceding layer $s$, used in
$W_P=B_{\ell-1}\cdots B_1$.
\\

\addlinespace[0.5em]
\multicolumn{2}{@{}l@{}}{
    \textbf{Global upper-bound proof}
}\\
\addlinespace[0.3em]

$\eta_{\ell,\diamond}$
&
The maximum number of distinct outgoing knots per source
in layer $\ell$:
$\eta_{\ell,\mathrm{ind}}=n_\ell K$
and $\eta_{\ell,\mathrm{sh}}=K$.
\\

$\mathscr P_{\ell,\diamond}$
&
The family of full-dimensional open convex cells in the
recursive subdivision after layer $\ell$, starting from
$\mathscr P_{0,\diamond}=\{\R^d\}$.
\\

$\mathscr Q_{\ell,\diamond}(P)$
&
The child cells obtained by dividing a parent cell $P$
using the proper knot preimages in layer $\ell$.
\\

$N_{\ell,\diamond}$
&
The number of cells in $\mathscr P_{\ell,\diamond}$,
with $N_{0,\diamond}=1$.
\\

$\mathcal H_{\diamond}^{\mathrm{cut}}$
&
The finite collection of ambient hyperplanes containing
the proper knot cuts used in the recursive subdivision.
\\

$L_U$, $L_P$
&
The affine extensions of $f$ from a linear region $U$
and a final subdivision cell $P$, respectively.
\\

$x_0$
&
A point in the nonempty open intersection $U\cap P$,
used to compare the affine maps $L_U$ and $L_P$.
\\

\addlinespace[0.5em]
\multicolumn{2}{@{}l@{}}{
    \textbf{Lower-bound construction: coordinate groups and folds}
}\\
\addlinespace[0.3em]

$\pi_\ell$
&
The surjection from active source indices to active
destination indices,
$\pi_\ell:[\rho_{\ell-1}]\to[\rho_\ell]$,
defined by $\pi_\ell(i)=\min\{i,\rho_\ell\}$.
\\

$G_{\ell,j}$
&
The nonempty group
$G_{\ell,j}=\pi_\ell^{-1}(\{j\})$
of active sources assigned to destination $j$.
These groups partition $[\rho_{\ell-1}]$.
\\

$\mathfrak p_{\ell,i}$
&
A prime assigned to active source $(\ell,i)$.
The primes are pairwise distinct across all active
layer/source pairs.
\\

$r$
&
In the folding construction, a branch index in
$\{0,\ldots,k-1\}$.
\\

$\varepsilon_r$
&
The alternating sign $\varepsilon_r=(-1)^r$.
\\

$\zeta_{\ell,i}$
&
The positive normalizing constant
$\zeta_{\ell,i}
=\sum_{u=0}^{k-1}\mathfrak p_{\ell,i}^{\,u}$.
\\

$\lambda_{\ell,i,r}$
&
The length of branch $r$,
$\lambda_{\ell,i,r}
=\mathfrak p_{\ell,i}^{\,r}/\zeta_{\ell,i}$.
These lengths are positive and sum to $1$.
\\

$\xi_{\ell,i,r}$
&
The branch endpoints, with $\xi_{\ell,i,0}=0$ and
$\xi_{\ell,i,r}
=\sum_{u=0}^{r-1}\lambda_{\ell,i,u}$
for $r\in[k]$. In particular, $\xi_{\ell,i,k}=1$.
\\

$\gamma_{\ell,i,r}$
&
The affine formula of branch $r$, with slope
$\varepsilon_r/\lambda_{\ell,i,r}$.
Its values at the two branch endpoints are
$(1-\varepsilon_r)/2$ and $(1+\varepsilon_r)/2$.
\\

$g_{\ell,i}$
&
The scalar $k$-segment folding function assigned to
active source $(\ell,i)$, obtained from the formulas
$\gamma_{\ell,i,r}$ and affine extensions at both ends.
\\

$g'_{\ell,i,-}$,
$g'_{\ell,i,+}$
&
The left and right derivatives of $g_{\ell,i}$,
respectively.
\\

$I_{\ell,i,r}$
&
The branch interval
$(\xi_{\ell,i,r},\xi_{\ell,i,r+1})$.
The restriction of $g_{\ell,i}$ to this interval
is an affine bijection onto $(0,1)$.
\\

$\widehat F_\ell$
&
The active coordinates of the partial network:
$\widehat F_\ell
=(z_1^{(\ell)},\ldots,z_{\rho_\ell}^{(\ell)})$.
Here $\widehat F_0$ is the identity and $\widehat F_T=f$.
\\

\addlinespace[0.5em]
\multicolumn{2}{@{}l@{}}{
    \textbf{Lower-bound construction: cells and distinct gradients}
}\\
\addlinespace[0.3em]

$\mathscr P_\ell$
&
The family of constructed cells inside $(0,1)^d$
after layer $\ell$. Every $P\in\mathscr P_\ell$
is mapped by $\widehat F_\ell$ onto
$(0,1)^{\rho_\ell}$.
\\

$\bm a$
&
A tuple of branch indices
$\bm a=(a_1,\ldots,a_{\rho_{\ell-1}})
\in\{0,\ldots,k-1\}^{\rho_{\ell-1}}$
for layer $\ell$.
\\

$B_{\ell,\bm a}$
&
The branch box
$\prod_{i=1}^{\rho_{\ell-1}}I_{\ell,i,a_i}$
in the active representation before layer $\ell$.
\\

$P_{\bm a}$
&
The child cell
$P\cap\widehat F_{\ell-1}^{-1}(B_{\ell,\bm a})$
of a fixed parent $P\in\mathscr P_{\ell-1}$.
\\

$y$, $u_i$, $t_i$
&
In the image-surjectivity argument, an arbitrary target
$y\in(0,1)^{\rho_\ell}$, the value
$u_i=y_{\pi_\ell(i)}$, and its unique preimage
$t_i\in I_{\ell,i,a_i}$ under the selected branch
of $g_{\ell,i}$.
\\

$\alpha$, $\beta$
&
Arrays of selected branch indices at all active sources,
with entries
$\alpha_{\ell,i},\beta_{\ell,i}\in\{0,\ldots,k-1\}$.
\\

$P_\alpha$, $P_\beta$
&
Final constructed cells associated with the branch arrays
$\alpha$ and $\beta$, respectively.
\\

$h$, $i_\ell(h)$
&
An input-coordinate index $h\in[d]$ and the active node
reached from that coordinate after layer $\ell$.
They satisfy $i_0(h)=h$ and
$i_\ell(h)=\pi_\ell(i_{\ell-1}(h))$.
\\

$(\ell_*,i_*)$
&
An active layer/source pair at which two branch arrays
differ:
$\alpha_{\ell_*,i_*}\neq\beta_{\ell_*,i_*}$.
\\

$L_\alpha$
&
The affine extension to $\R^d$ of the restriction
$f|_{P_\alpha}$.
\\

$\Omega_\alpha$
&
The open set
$\operatorname{int}
\{x\in\R^d:f(x)=L_\alpha(x)\}$.
\\

$U_\alpha$
&
The connected component of $\Omega_\alpha$ containing
$P_\alpha$. It is a linear region of $f$.
\\

$V$, $L_V$
&
A connected open affine set considered in the maximality
argument and its affine formula, respectively.
\\

\addlinespace[0.5em]
\multicolumn{2}{@{}l@{}}{
    \textbf{Existing KAN bounds and ReLU conversion}
}\\
\addlinespace[0.3em]

$n$
&
The common hidden-layer width when all KAN hidden
layers have equal width.
\\

$\displaystyle
\sum_{\ell=1}^{T}n_{\ell-1}n_\ell$
&
The total number of KAN edge functions and the exponent
in the upper bound of
Schoots et al.~\cite{schoots2025relating}.
\\

$c_{\diamond,\mathbf n}$,
$C_{\diamond,\mathbf n}$
&
Positive constants in the asymptotic comparisons.
They may depend on $\mathbf n$ and the model
$\diamond$ but are independent of $k$.
\\

$\sigma$
&
The ReLU function $\sigma(t)=\max\{0,t\}$.
\\

$s_{j,i}$
&
The number of actual knots of $\phi_{j,i}$,
with $s_{j,i}\leq K$.
\\

$\tau_{j,i,r}$
&
The $r$-th knot of $\phi_{j,i}$. In the construction
attaining the edge-wise single-layer bound, knots are
distinct across outgoing edges from the same source.
\\

$a_{j,i}$,
$b_{j,i,0}$
&
The constant and linear coefficients of the affine
part in the shifted-ReLU representation of
$\phi_{j,i}$.
\\

$b_{j,i,r}$,
$c_{j,i,r}$
&
The slope immediately after the $r$-th knot and its
slope change
$c_{j,i,r}=b_{j,i,r}-b_{j,i,r-1}$, respectively.
\\

$W$, $w_{j,i}$, $b_j$
&
The weight matrix, individual weights, and biases
of a ReLU layer.
\\

$y_j$
&
The ReLU output
$y_j=\sigma(\sum_i w_{j,i}z_i+b_j)$.
\\

$\Psi$, $\psi_{j,i}$
&
The first KAN layer and its affine edge functions
in the conversion of a ReLU layer into two KAN layers.
\\

$\Gamma$, $\gamma_{j,i}$
&
The second KAN layer and its edge functions, with
ReLU on diagonal edges and zero on other edges.
\\

\addlinespace[0.5em]
\multicolumn{2}{@{}l@{}}{
    \textbf{Comparison under a common parameter budget}
}\\
\addlinespace[0.3em]

$\Lambda$
&
The common upper bound on the total parameter count.
All parameter slots in the specified dense
parameterizations are counted, including zero entries.
\\

$\Lambda_{\mathrm{KAN}}(\mathbf n,k)$
&
The edge-wise-knot KAN parameter count
$2k\sum_{\ell=1}^{T}n_{\ell-1}n_\ell$.
Each edge has two affine coefficients, $K$ slope-change
coefficients, and $K$ trainable candidate knot locations.
\\

$\mathbf m$, $m_\ell$
&
The ReLU width vector
$\mathbf m=(m_0,\ldots,m_{T+1})$ and its layer widths,
with $m_0=d$, $m_{T+1}=1$, and $T$ ReLU hidden
layers followed by an affine scalar output layer.
\\

$\Lambda_{\mathrm{ReLU}}(\mathbf m)$
&
The ReLU parameter count
$\sum_{\ell=1}^{T+1}(m_{\ell-1}+1)m_\ell$,
including all weights and biases.
\\

$\mathcal R^\star_{\mathrm{KAN}}(d,T,\Lambda)$
&
The maximum number of linear regions among edge-wise-knot
CPWL KANs with input dimension $d$, $T$ layers, and at most
$\Lambda$ parameters. Hidden widths and the maximum number
of affine segments per edge $k$ may vary.
\\

$\mathcal R^\star_{\mathrm{ReLU}}(d,T,\Lambda)$
&
The maximum number of linear regions among ReLU networks
with input dimension $d$, $T$ ReLU hidden layers,
an affine scalar output, and parameter count at most
$\Lambda$. Hidden widths may vary.
\\

$\chi_d$
&
The constant $\chi_d=\sum_{j=0}^{d}1/j!$ used in the upper
bounds of Theorem~\ref{thm:parameter-budget-comparison}.
\\

$\nu_T$
&
The integer $\nu_T=\lceil T/2\rceil$, equal to the
number of odd-indexed hidden layers in the ReLU
lower-bound construction.
\\

$e_\ell$
&
The number of edges in KAN layer $\ell$:
$e_\ell=n_{\ell-1}n_\ell$.
\\

$e_{\mathrm{base}}$
&
The edge count
$e_{\mathrm{base}}=(T-1)d^2+d$
of the scalar-output KAN whose hidden widths all equal $d$.
\\

$k_\Lambda$
&
The maximum number of affine segments per edge chosen for
the KAN lower-bound construction:
$k_\Lambda=\lfloor\Lambda/(2e_{\mathrm{base}})\rfloor$.
The construction uses at most $\Lambda$ parameters.
\\

$\eta_{d,T}$
&
The constant $\eta_{d,T}=(T+1)(d+1)$
used to control the parameter count of the ReLU
lower-bound construction.
\\

$\mu_\Lambda$
&
The width
$\mu_\Lambda=\lfloor\Lambda/\eta_{d,T}\rfloor-1$.
The ReLU construction uses width $\mu_\Lambda$
at odd-indexed hidden layers and width $d$
at even-indexed hidden layers.
\\

$\mathbf h$
&
The final hidden representation
$\mathbf h:\R^d\to\R^{m_T}$
in the ReLU lower-bound construction.
\\

$\iota$
&
The number of distinct affine formulas of $\mathbf h$.
\\

$D_a$, $\delta_a$
&
The coefficients of the $a$-th distinct affine formula
$\mathbf h(x)=D_ax+\delta_a$, where
$D_a\in\R^{m_T\times d}$,
$\delta_a\in\R^{m_T}$, and $a\in[\iota]$.
\\

$\omega$
&
An output weight vector in $\R^{m_T}$ chosen so that
$\omega^{\mathsf T}\mathbf h$ preserves the linear
regions of $\mathbf h$.
\\

$\omega'$
&
A candidate vector in $\R^{m_T}$ used to describe
the exceptional output weights that identify
two distinct affine formulas.
\\

\addlinespace[0.5em]
\multicolumn{2}{@{}l@{}}{
    \textbf{Parallel-family arrangement proof}
}\\
\addlinespace[0.3em]

$\mathcal A$, $\mathcal B$, $\mathcal D$
&
Hyperplane arrangements. In the induction,
$\mathcal A$ is the full arrangement and $\mathcal B$
contains all families except the last one.
\\

$\mathscr C(\mathcal D)$,
$N(\mathcal D)$
&
The set of regions of $\mathcal D$ and its cardinality.
\\

$C$ in $\mathscr C(\mathcal D)$
&
An open convex region of the arrangement $\mathcal D$.
\\

$G$, $H$
&
Individual hyperplanes. In the insertion argument,
$H$ is the newly inserted hyperplane.
\\

$\mathcal D|_H$
&
The arrangement on $H$ formed by the nonempty
intersections $G\cap H$ for $G\in\mathcal D$,
where $H\notin\mathcal D$.
\\

$\mathscr C_H(\mathcal D)$
&
The regions of $\mathcal D$ that intersect $H$.
\\

$t$, $H_1,\ldots,H_t$
&
The number and members of the last parallel family,
with $0\leq t\leq\kappa$.
\\

$\mathcal A_a$
&
The intermediate arrangement
$\mathcal B\cup\{H_1,\ldots,H_a\}$,
with $\mathcal A_0=\mathcal B$.
\\

$\mathcal J$
&
A subcollection of $\mathcal B$ used to verify
general position of the arrangement induced on $H$.
\\

\end{longtable}
\endgroup

\subsection{Proof of Proposition~\ref{prop:parallel-arrangement}}

\begin{proof}
For an arrangement $\mathcal D$, let $\mathscr C(\mathcal D)$
be its set of regions and let
$N(\mathcal D):=\abs{\mathscr C(\mathcal D)}$.
Use the conventions
\[
    A_r(0,\kappa)=A_0(m,\kappa)=1,
    \quad
    \binom ab=0
    \quad\text{for }b<0\text{ or }b>a.
\]

Let $H\notin\mathcal D$ be a hyperplane, and define
\[
\begin{aligned}
    \mathcal D|_H
    &:=
    \{G\cap H:G\in\mathcal D,\ G\cap H\neq\varnothing\},\\
    \mathscr C_H(\mathcal D)
    &:=
    \{C\in\mathscr C(\mathcal D):C\cap H\neq\varnothing\}.
\end{aligned}
\]
Regions of $\mathcal D|_H$ are taken relative to $H$.
Since every $C\in\mathscr C(\mathcal D)$ is open and convex,
\[
    C\longmapsto C\cap H
\]
is a bijection from $\mathscr C_H(\mathcal D)$ to
$\mathscr C(\mathcal D|_H)$. Inserting $H$ splits each
$C\in\mathscr C_H(\mathcal D)$ into two nonempty convex
regions. Hence
\[
\begin{aligned}
    N(\mathcal D\cup\{H\})
    &=
    N(\mathcal D)
    -\abs{\mathscr C_H(\mathcal D)}
    +2\abs{\mathscr C_H(\mathcal D)}\\
    &=
    N(\mathcal D)+\abs{\mathscr C_H(\mathcal D)}\\
    &=
    N(\mathcal D)+N(\mathcal D|_H).
\end{aligned}
\]

We prove the bound by induction on $m$, simultaneously for
all $r\geq0$. For $m=0$ or $r=0$, the region count is one.
Assume the bound holds for at most $m-1$ families and let
$r\geq1$.

After removing repeated hyperplanes, write
\[
    \mathcal A
    =
    \mathcal B\cup\{H_1,\ldots,H_t\},
    \quad
    0\leq t\leq\kappa,
\]
where $\mathcal B$ contains at most $m-1$ families and
$H_1,\ldots,H_t$ are mutually parallel. Set
\[
    \mathcal A_0:=\mathcal B,
    \quad
    \mathcal A_a
    :=
    \mathcal B\cup\{H_1,\ldots,H_a\},
    \quad a\in[t].
\]
Parallelism gives
\[
    \mathcal A_{a-1}|_{H_a}
    =
    \mathcal B|_{H_a}.
\]
The induced arrangement $\mathcal B|_{H_a}$ lies in dimension
$r-1$ and consists of at most $m-1$ parallel families, each
containing at most $\kappa$ hyperplanes. Thus
\[
\begin{aligned}
    N(\mathcal B)
    &\leq A_r(m-1,\kappa),\\
    N(\mathcal B|_{H_a})
    &\leq A_{r-1}(m-1,\kappa).
\end{aligned}
\]
Therefore,
\[
\begin{aligned}
    N(\mathcal A)
    &=
    N(\mathcal A_0)
    +
    \sum_{a=1}^{t}
    \bigl[N(\mathcal A_a)-N(\mathcal A_{a-1})\bigr]\\
    &=
    N(\mathcal B)
    +
    \sum_{a=1}^{t}N(\mathcal B|_{H_a})\\
    &\leq
    A_r(m-1,\kappa)
    +
    tA_{r-1}(m-1,\kappa)\\
    &\leq
    A_r(m-1,\kappa)
    +
    \kappa A_{r-1}(m-1,\kappa)\\
    &=
    \sum_{s=0}^{r}
    \binom{m-1}{s}\kappa^s
    +
    \sum_{s=0}^{r-1}
    \binom{m-1}{s}\kappa^{s+1}\\
    &=
    \sum_{s=0}^{r}
    \binom{m-1}{s}\kappa^s
    +
    \sum_{s=1}^{r}
    \binom{m-1}{s-1}\kappa^s\\
    &=
    \sum_{s=0}^{r}
    \left[
        \binom{m-1}{s}
        +
        \binom{m-1}{s-1}
    \right]\kappa^s\\
    &=
    \sum_{s=0}^{r}\binom ms\kappa^s\\
    &=
    \sum_{s=0}^{\min\{r,m\}}\binom ms\kappa^s\\
    &=
    A_r(m,\kappa).
\end{aligned}
\]

For $\kappa=1$, let $\mathcal A$ consist of $m$ hyperplanes
in general position. Choose $H\in\mathcal A$ and put
$\mathcal B:=\mathcal A\setminus\{H\}$.
For every $\mathcal J\subseteq\mathcal B$ with
$1\leq\abs{\mathcal J}\leq\min\{r-1,m-1\}$,
\[
\begin{aligned}
    \dim\left(
        H\cap\bigcap_{G\in\mathcal J}G
    \right)
    &=
    r-\bigl(\abs{\mathcal J}+1\bigr)\\
    &=
    (r-1)-\abs{\mathcal J}.
\end{aligned}
\]
For every $\mathcal J\subseteq\mathcal B$ with
$\abs{\mathcal J}=r$,
\[
    H\cap\bigcap_{G\in\mathcal J}G
    =
    \varnothing.
\]
Thus, for $r\geq2$, $\mathcal B|_H$ consists of $m-1$
hyperplanes in general position in $H$. For $r=1$, it is
the empty arrangement on the point $H$.

Induction on $m$, starting from the empty arrangement, gives
\[
\begin{aligned}
    N(\mathcal A)
    &=
    N(\mathcal B)+N(\mathcal B|_H)\\
    &=
    A_r(m-1,1)+A_{r-1}(m-1,1)\\
    &=
    \sum_{s=0}^{r}\binom{m-1}{s}
    +
    \sum_{s=0}^{r-1}\binom{m-1}{s}\\
    &=
    \sum_{s=0}^{r}\binom{m-1}{s}
    +
    \sum_{s=1}^{r}\binom{m-1}{s-1}\\
    &=
    \sum_{s=0}^{r}
    \left[
        \binom{m-1}{s}
        +
        \binom{m-1}{s-1}
    \right]\\
    &=
    \sum_{s=0}^{r}\binom ms\\
    &=
    \sum_{s=0}^{\min\{r,m\}}\binom ms\\
    &=
    A_r(m,1).
\end{aligned}
\]
This completes the proof.
\end{proof}

\subsection{Proof of Proposition~\ref{prop:one-layer-kan}}

\begin{proof}
For each $i\in[p]$, write
\[
    \mathcal K_i^{\mathrm{out}}
    =
    \{\tau_{i,1}<\cdots<\tau_{i,M_i}\},
\]
and set
\[
    \tau_{i,0}:=-\infty,
    \quad
    \tau_{i,M_i+1}:=+\infty.
\]
Define
\[
    I_{i,r}:=(\tau_{i,r},\tau_{i,r+1}),
    \quad r\in\{0,\ldots,M_i\}.
\]
For
$\bm r=(r_1,\ldots,r_p)
\in\prod_{i=1}^{p}\{0,\ldots,M_i\}$, let
\[
    B_{\bm r}:=\prod_{i=1}^{p}I_{i,r_i}.
\]
Every edge function is affine on the corresponding interval:
\[
    \phi_{j,i}(t)
    =
    a_{j,i,r_i}t+b_{j,i,r_i},
    \quad t\in I_{i,r_i},
\]
for some $a_{j,i,r_i},b_{j,i,r_i}\in\R$. Hence, on $B_{\bm r}$,
\[
\begin{aligned}
    [\Phi(z)]_j
    &=
    \sum_{i=1}^{p}\phi_{j,i}(z_i)\\
    &=
    \sum_{i=1}^{p}
    \bigl(a_{j,i,r_i}z_i+b_{j,i,r_i}\bigr)\\
    &=
    \sum_{i=1}^{p}a_{j,i,r_i}z_i
    +
    \sum_{i=1}^{p}b_{j,i,r_i}.
\end{aligned}
\]
Thus $\Phi$ is affine on every $B_{\bm r}$.

Define
\[
    \mathcal H
    :=
    \bigcup_{i=1}^{p}
    \bigcup_{\tau\in\mathcal K_i^{\mathrm{out}}}
    \{z\in\R^p:z_i=\tau\}.
\]
Fix $z\in\mathcal H$, and choose $i$ and $\tau=z_i$ such that
$\tau\in\mathcal K_i^{\mathrm{out}}$. There exists $j\in[q]$ with
$\tau\in\mathcal K(\phi_{j,i})$. Let $a^-$ and $a^+$ denote the
left and right slopes of $\phi_{j,i}$ at $\tau$. Then
\[
    a^-\neq a^+.
\]
Writing $e_i\in\R^p$ for the $i$th coordinate vector gives
\[
\begin{aligned}
    \lim_{h\downarrow0}
    \frac{[\Phi(z+he_i)]_j-[\Phi(z)]_j}{h}
    &=
    \lim_{h\downarrow0}
    \frac{\phi_{j,i}(\tau+h)-\phi_{j,i}(\tau)}{h}\\
    &=a^+,
\end{aligned}
\]
and
\[
\begin{aligned}
    \lim_{h\downarrow0}
    \frac{[\Phi(z)]_j-[\Phi(z-he_i)]_j}{h}
    &=
    \lim_{h\downarrow0}
    \frac{\phi_{j,i}(\tau)-\phi_{j,i}(\tau-h)}{h}\\
    &=a^-.
\end{aligned}
\]
Therefore $\Phi$ is not affine on any neighborhood of $z$.

Let $U$ be a connected open set on which $\Phi$ is affine.
The preceding argument gives
\[
    U\cap\mathcal H=\varnothing.
\]
For each $i$, the coordinate projection
$\{z_i:z\in U\}$ is connected and avoids
$\mathcal K_i^{\mathrm{out}}$. Consequently, it is contained in
one interval $I_{i,r_i}$, and therefore
\[
    U\subseteq\prod_{i=1}^{p}I_{i,r_i}=B_{\bm r}.
\]
Since $B_{\bm r}$ is connected, open, and affine for $\Phi$,
the maximal connected open affine regions are exactly the boxes
$B_{\bm r}$. Thus
\[
    \mathcal R(\Phi)
    =
    \prod_{i=1}^{p}(M_i+1).
\]

In the edge-wise-knot model,
\[
\begin{aligned}
    M_i
    &=
    \left|
        \bigcup_{j=1}^{q}\mathcal K(\phi_{j,i})
    \right|\\
    &\leq
    \sum_{j=1}^{q}|\mathcal K(\phi_{j,i})|\\
    &\leq qK.
\end{aligned}
\]
Hence
\[
\begin{aligned}
    \mathcal R(\Phi)
    &=
    \prod_{i=1}^{p}(M_i+1)\\
    &\leq(qK+1)^p\\
    &=
    \bigl(q(k-1)+1\bigr)^p.
\end{aligned}
\]
For each $i$, choose $qK$ pairwise distinct real numbers
$\tau_{j,i,r}$, indexed by $j\in[q]$ and $r\in[K]$, and define
\[
    \phi_{j,i}(t)
    :=
    \sum_{r=1}^{K}\sigma(t-\tau_{j,i,r}).
\]
Each slope jump equals $1$, so
\[
    \mathcal K(\phi_{j,i})
    =
    \{\tau_{j,i,r}:r\in[K]\},
    \quad
    M_i=qK.
\]
This attains $(qK+1)^p$ regions.

In the source-shared-knot model,
\[
    \mathcal K_i^{\mathrm{out}}
    \subseteq\mathcal S_i,
    \quad
    |\mathcal S_i|\leq K.
\]
Therefore
\[
\begin{aligned}
    \mathcal R(\Phi)
    &=
    \prod_{i=1}^{p}(M_i+1)\\
    &\leq(K+1)^p\\
    &=k^p.
\end{aligned}
\]
To attain equality, choose $|\mathcal S_i|=K$ and set
\[
    \phi_{1,i}(t)
    :=
    \sum_{\tau\in\mathcal S_i}\sigma(t-\tau),
    \quad
    \phi_{j,i}(t):=0
    \quad\text{for }j\geq2.
\]
Then $\mathcal K_i^{\mathrm{out}}=\mathcal S_i$, so
\[
    \mathcal R(\Phi)
    =
    \prod_{i=1}^{p}(K+1)
    =
    k^p.
\]
This completes the proof.
\end{proof}

\subsection{Proof of Lemma~\ref{lem:layerwise-subdivision}}

\begin{proof}
Write the $i$th row of $W_P$ as $w_{P,i}^{\mathsf T}$.
For $x\in P$,
\[
    [F_{\ell-1}(x)]_i
    =
    w_{P,i}^{\mathsf T}x+(c_P)_i.
\]
A knot $\tau$ associated with source coordinate $i$ has preimage
\[
    H_{P,i,\tau}
    :=
    \left\{
        x\in P:
        w_{P,i}^{\mathsf T}x+(c_P)_i=\tau
    \right\}.
\]

If $w_{P,i}=0$, then
\[
    H_{P,i,\tau}
    =
    \begin{cases}
        P,&(c_P)_i=\tau,\\
        \varnothing,&(c_P)_i\neq\tau.
    \end{cases}
\]
In either case, the $i$th source coordinate is constant on $P$
and creates no subdivision. All such rows may therefore be
omitted from the arrangement.

If $r_P=0$, then $W_P=0$ and
\[
    F_\ell(x)=\Phi_\ell(c_P),
    \quad x\in P.
\]
Thus there is one cell, in agreement with
\[
    A_0(n_{\ell-1},n_\ell K)
    =
    A_0(n_{\ell-1},K)
    =
    1.
\]

Suppose $r_P>0$. Choose a rank factorization
\[
    W_P=V_PL_P,
\]
where
\[
    V_P\in\R^{n_{\ell-1}\times r_P},
    \quad
    L_P\in\R^{r_P\times d},
    \quad
    \operatorname{rank}(L_P)=r_P.
\]
Let $v_{P,i}^{\mathsf T}$ be the $i$th row of $V_P$.
For every nonzero row $w_{P,i}$, the corresponding cuts become,
in the coordinate $y=L_Px$,
\[
    v_{P,i}^{\mathsf T}y+(c_P)_i=\tau.
\]
For a fixed $i$, these are parallel hyperplanes in $\R^{r_P}$.

Let $D$ be an open region of this arrangement. Its preimage in
$P$ is
\[
    Q_D:=P\cap L_P^{-1}(D).
\]
Both $P$ and $D$ are convex, so $Q_D$ is convex. A nonempty
$Q_D$ is open and connected. Consequently, each arrangement
region contributes at most one child cell:
\[
    \#\{\text{child cells in }P\}
    \leq
    \#\{\text{arrangement regions in }\R^{r_P}\}.
\]

On each nonempty $Q_D$, every nonconstant source coordinate
stays in one knot interval. A constant source coordinate
contributes a constant edge value, including when it equals a
knot. Hence there exist
\[
    D_{Q_D}\in\R^{n_\ell\times n_{\ell-1}},
    \quad
    b_{Q_D}\in\R^{n_\ell}
\]
such that
\[
\begin{aligned}
    F_\ell(x)
    &=
    D_{Q_D}F_{\ell-1}(x)+b_{Q_D}\\
    &=
    D_{Q_D}(W_Px+c_P)+b_{Q_D}\\
    &=
    D_{Q_D}W_Px+
    D_{Q_D}c_P+b_{Q_D},
    \quad x\in Q_D.
\end{aligned}
\]
Thus every child cell is an affine cell for $F_\ell$.

In the edge-wise-knot model, each source contributes at most
\[
\begin{aligned}
    \left|
        \bigcup_{j=1}^{n_\ell}
        \mathcal K(\phi_{\ell,j,i})
    \right|
    &\leq
    \sum_{j=1}^{n_\ell}
    |\mathcal K(\phi_{\ell,j,i})|\\
    &\leq n_\ell K
\end{aligned}
\]
distinct knots. The parallel-family arrangement bound yields
\[
\begin{aligned}
    \#\{\text{child cells in }P\}
    &\leq
    A_{r_P}(n_{\ell-1},n_\ell K)\\
    &=
    A_{r_P}\bigl(n_{\ell-1},n_\ell(k-1)\bigr).
\end{aligned}
\]

In the source-shared-knot model,
\[
    \bigcup_{j=1}^{n_\ell}
    \mathcal K(\phi_{\ell,j,i})
    \subseteq\mathcal S_{\ell,i},
    \quad
    |\mathcal S_{\ell,i}|\leq K.
\]
Therefore
\[
\begin{aligned}
    \#\{\text{child cells in }P\}
    &\leq
    A_{r_P}(n_{\ell-1},K)\\
    &=
    A_{r_P}(n_{\ell-1},k-1).
\end{aligned}
\]

It remains to bound $r_P$. For $\ell=1$,
\[
    W_P=I_d,
    \quad
    r_P=d=\rho_0.
\]
For $\ell\geq2$, successively subdividing by the proper knot
preimages gives a finite convex polyhedral refinement for the
first $\ell-1$ layers. Choose one cell $C$ of this refinement
such that $C\cap P$ is nonempty and open.

On $C$, there are matrices and vectors
\[
    B_s\in\R^{n_s\times n_{s-1}},
    \quad
    b_s\in\R^{n_s},
    \quad s\in[\ell-1],
\]
satisfying
\[
    F_s(x)=B_sF_{s-1}(x)+b_s.
\]
Differentiating these affine identities on $C\cap P$ gives
\[
\begin{aligned}
    W_P
    &=
    D F_{\ell-1}\\
    &=
    B_{\ell-1}D F_{\ell-2}\\
    &=
    B_{\ell-1}\cdots B_1.
\end{aligned}
\]
Thus
\[
\begin{aligned}
    r_P
    &=
    \operatorname{rank}(B_{\ell-1}\cdots B_1)\\
    &\leq
    \min_{s\in[\ell-1]}\operatorname{rank}(B_s)\\
    &\leq
    \min\{n_0,\ldots,n_{\ell-1}\}\\
    &=
    \rho_{\ell-1}.
\end{aligned}
\]
This completes the proof.
\end{proof}

\subsection{Proof of Theorem~\ref{thm:global-upper}}

\begin{proof}
Fix $\diamond\in\{\mathrm{ind},\mathrm{sh}\}$ and
$f\in\mathcal F_\diamond(\mathbf n,k)$. Define
\[
    \eta_{\ell,\diamond}
    :=
    \begin{cases}
        n_\ell K,&\diamond=\mathrm{ind},\\
        K,&\diamond=\mathrm{sh},
    \end{cases}
    \quad \ell\in[T].
\]
Set
\[
    \mathscr P_{0,\diamond}:=\{\R^d\},
    \quad
    N_{0,\diamond}:=1.
\]
For $P\in\mathscr P_{\ell-1,\diamond}$, let
$\mathscr Q_{\ell,\diamond}(P)$ be the nonempty open cells
obtained by cutting $P$ with the proper knot preimages
of layer $\ell$. Define
\[
    \mathscr P_{\ell,\diamond}
    :=
    \bigcup_{P\in\mathscr P_{\ell-1,\diamond}}
    \mathscr Q_{\ell,\diamond}(P),
    \quad
    N_{\ell,\diamond}
    :=
    |\mathscr P_{\ell,\diamond}|.
\]
All cells are convex, and distinct cells are disjoint.

On $P\in\mathscr P_{\ell-1,\diamond}$, write
\[
    F_{\ell-1}(x)=W_Px+c_P,
    \quad
    r_P:=\operatorname{rank}(W_P).
\]
Constant source coordinates require no cuts, since
\[
    (W_P)_{i,:}=0
    \quad\Longrightarrow\quad
    \phi_{\ell,j,i}\!\left([F_{\ell-1}(x)]_i\right)
    =
    \phi_{\ell,j,i}\!\left((c_P)_i\right).
\]
Every nonconstant source coordinate remains between
consecutive knots on each child cell. Hence $F_\ell$
is affine on every cell of $\mathscr P_{\ell,\diamond}$.

For $\kappa\in\mathbb N$ and integers
$0\leq r\leq r'\leq m$,
\[
\begin{aligned}
    A_{r'}(m,\kappa)-A_r(m,\kappa)
    &=
    \sum_{s=0}^{r'}\binom ms\kappa^s
    -
    \sum_{s=0}^{r}\binom ms\kappa^s\\
    &=
    \sum_{s=r+1}^{r'}\binom ms\kappa^s\\
    &\geq0.
\end{aligned}
\]
By Lemma~\ref{lem:layerwise-subdivision},
\[
    r_P\leq\rho_{\ell-1}\leq n_{\ell-1},
\]
and therefore
\[
\begin{aligned}
    N_{\ell,\diamond}
    &=
    \sum_{P\in\mathscr P_{\ell-1,\diamond}}
    |\mathscr Q_{\ell,\diamond}(P)|\\
    &\leq
    \sum_{P\in\mathscr P_{\ell-1,\diamond}}
    A_{r_P}
    (n_{\ell-1},\eta_{\ell,\diamond})\\
    &\leq
    \sum_{P\in\mathscr P_{\ell-1,\diamond}}
    A_{\rho_{\ell-1}}
    (n_{\ell-1},\eta_{\ell,\diamond})\\
    &=
    N_{\ell-1,\diamond}
    A_{\rho_{\ell-1}}
    (n_{\ell-1},\eta_{\ell,\diamond}).
\end{aligned}
\]
Starting from $N_{0,\diamond}=1$, induction gives
\[
\begin{aligned}
    N_{t,\diamond}
    &\leq
    N_{t-1,\diamond}
    A_{\rho_{t-1}}
    (n_{t-1},\eta_{t,\diamond})\\
    &\leq
    \left[
        \prod_{\ell=1}^{t-1}
        A_{\rho_{\ell-1}}
        (n_{\ell-1},\eta_{\ell,\diamond})
    \right]
    A_{\rho_{t-1}}
    (n_{t-1},\eta_{t,\diamond})\\
    &=
    \prod_{\ell=1}^{t}
    A_{\rho_{\ell-1}}
    (n_{\ell-1},\eta_{\ell,\diamond}),
    \quad t\in[T].
\end{aligned}
\]

Let $\mathcal H^{\mathrm{cut}}_\diamond$ be the finite
collection of ambient hyperplanes used for the proper cuts.
Then
\[
    \R^d\setminus
    \bigcup_{P\in\mathscr P_{T,\diamond}}P
    \subseteq
    \bigcup_{H\in\mathcal H^{\mathrm{cut}}_\diamond}H,
    \quad
    \operatorname{int}
    \left(
        \bigcup_{H\in\mathcal H^{\mathrm{cut}}_\diamond}H
    \right)
    =
    \varnothing.
\]
Thus every nonempty open linear region $U$ of $f$
intersects some $P\in\mathscr P_{T,\diamond}$.

Let $L_U$ and $L_P$ be the affine formulas of $f$ on
$U$ and $P$, respectively. Since $U\cap P$ is nonempty
and open,
\[
    L_U=L_P
    \quad\text{on }U\cap P,
    \quad
    \nabla L_U=\nabla L_P.
\]
For $x_0\in U\cap P$ and every $x\in\R^d$,
\[
\begin{aligned}
    L_U(x)-L_P(x)
    &=
    L_U(x_0)-L_P(x_0)
    +
    (\nabla L_U-\nabla L_P)^{\mathsf T}(x-x_0)\\
    &=0.
\end{aligned}
\]
Consequently, $U\cup P$ is a connected open set on which
$f=L_U$. Maximality of $U$ gives
\[
    U\cup P=U,
    \quad P\subseteq U.
\]
If two linear regions contain the same cell, their affine
formulas coincide and their union is connected and open;
maximality then makes the two regions equal. Hence
\[
    \mathcal R(f)
    \leq
    |\mathscr P_{T,\diamond}|
    =
    N_{T,\diamond}.
\]

Combining the preceding bounds,
\[
\begin{aligned}
    \mathcal R(f)
    &\leq
    \prod_{\ell=1}^{T}
    A_{\rho_{\ell-1}}
    (n_{\ell-1},\eta_{\ell,\diamond})\\
    &=
    \prod_{\ell=1}^{T}
    \left[
        \sum_{s=0}^{\rho_{\ell-1}}
        \binom{n_{\ell-1}}s
        \eta_{\ell,\diamond}^{\,s}
    \right].
\end{aligned}
\]
The attainable region counts form a nonempty bounded subset
of $\mathbb N$, so the maximum exists. Substituting
$K=k-1$ and the two values of $\eta_{\ell,\diamond}$ yields
\[
\begin{aligned}
    \mathcal R^\star_{\mathrm{ind}}(\mathbf n,k)
    &\leq
    \prod_{\ell=1}^{T}
    \left[
        \sum_{s=0}^{\rho_{\ell-1}}
        \binom{n_{\ell-1}}s
        \bigl(n_\ell(k-1)\bigr)^s
    \right],\\
    \mathcal R^\star_{\mathrm{sh}}(\mathbf n,k)
    &\leq
    \prod_{\ell=1}^{T}
    \left[
        \sum_{s=0}^{\rho_{\ell-1}}
        \binom{n_{\ell-1}}s(k-1)^s
    \right].
\end{aligned}
\]

For $0\leq s\leq\rho_{\ell-1}$ and $k\geq2$,
\[
    (k-1)^s
    \leq k^s
    \leq k^{\rho_{\ell-1}}.
\]
Therefore,
\[
\begin{aligned}
    &\sum_{s=0}^{\rho_{\ell-1}}
    \binom{n_{\ell-1}}s
    \bigl(n_\ell(k-1)\bigr)^s\\
    &\quad=
    \sum_{s=0}^{\rho_{\ell-1}}
    \binom{n_{\ell-1}}s n_\ell^s(k-1)^s\\
    &\quad\leq
    \sum_{s=0}^{\rho_{\ell-1}}
    \binom{n_{\ell-1}}s n_\ell^s k^{\rho_{\ell-1}}\\
    &\quad=
    k^{\rho_{\ell-1}}
    \sum_{s=0}^{\rho_{\ell-1}}
    \binom{n_{\ell-1}}s n_\ell^s,
\end{aligned}
\]
and
\[
\begin{aligned}
    \sum_{s=0}^{\rho_{\ell-1}}
    \binom{n_{\ell-1}}s(k-1)^s
    &\leq
    \sum_{s=0}^{\rho_{\ell-1}}
    \binom{n_{\ell-1}}s k^{\rho_{\ell-1}}\\
    &=
    k^{\rho_{\ell-1}}
    \sum_{s=0}^{\rho_{\ell-1}}
    \binom{n_{\ell-1}}s.
\end{aligned}
\]
Multiplying over $\ell\in[T]$ gives
\[
\begin{aligned}
    \mathcal R^\star_{\mathrm{ind}}(\mathbf n,k)
    &\leq
    \prod_{\ell=1}^{T}
    \left\{
        k^{\rho_{\ell-1}}
        \sum_{s=0}^{\rho_{\ell-1}}
        \binom{n_{\ell-1}}s n_\ell^s
    \right\}\\
    &=
    \left\{
        \prod_{\ell=1}^{T}
        \left[
            \sum_{s=0}^{\rho_{\ell-1}}
            \binom{n_{\ell-1}}s n_\ell^s
        \right]
    \right\}
    k^{\sum_{\ell=1}^{T}\rho_{\ell-1}},\\
    \mathcal R^\star_{\mathrm{sh}}(\mathbf n,k)
    &\leq
    \prod_{\ell=1}^{T}
    \left\{
        k^{\rho_{\ell-1}}
        \sum_{s=0}^{\rho_{\ell-1}}
        \binom{n_{\ell-1}}s
    \right\}\\
    &=
    \left\{
        \prod_{\ell=1}^{T}
        \left[
            \sum_{s=0}^{\rho_{\ell-1}}
            \binom{n_{\ell-1}}s
        \right]
    \right\}
    k^{\sum_{\ell=1}^{T}\rho_{\ell-1}}.
\end{aligned}
\]
For fixed $\mathbf n$, both coefficient products are
independent of $k$. Hence
\[
\begin{aligned}
    \mathcal R^\star_{\mathrm{ind}}(\mathbf n,k)
    &=
    O_{\mathbf n}\!\left(
        k^{\sum_{\ell=1}^{T}\rho_{\ell-1}}
    \right),\\
    \mathcal R^\star_{\mathrm{sh}}(\mathbf n,k)
    &=
    O_{\mathbf n}\!\left(
        k^{\sum_{\ell=1}^{T}\rho_{\ell-1}}
    \right).
\end{aligned}
\]
This completes the proof.
\end{proof}

\subsection{Proof of Theorem~\ref{thm:global-lower}}

\begin{proof}
Set $\rho_T:=1$. Then
\[
    \rho_0=d,
    \quad
    \rho_\ell=\min\{\rho_{\ell-1},n_\ell\},
    \quad \ell\in[T].
\]
For $\ell\in[T]$, define
\[
    \pi_\ell:[\rho_{\ell-1}]
    \longrightarrow[\rho_\ell],
    \quad
    \pi_\ell(i):=\min\{i,\rho_\ell\},
\]
and
\[
    G_{\ell,j}:=\pi_\ell^{-1}(\{j\}),
    \quad j\in[\rho_\ell].
\]
Since $\pi_\ell(j)=j$ for $j\in[\rho_\ell]$,
\[
    G_{\ell,j}\neq\varnothing,
    \quad
    \bigsqcup_{j=1}^{\rho_\ell}G_{\ell,j}
    =
    [\rho_{\ell-1}].
\]

Choose pairwise distinct primes $\mathfrak p_{\ell,i}$
for $\ell\in[T]$ and $i\in[\rho_{\ell-1}]$. Define
\[
    \varepsilon_r:=(-1)^r,
    \quad
    \zeta_{\ell,i}
    :=
    \sum_{u=0}^{k-1}\mathfrak p_{\ell,i}^{\,u},
    \quad
    \lambda_{\ell,i,r}
    :=
    \frac{\mathfrak p_{\ell,i}^{\,r}}{\zeta_{\ell,i}},
    \quad r\in\{0,\ldots,k-1\}.
\]
Then
\[
    \lambda_{\ell,i,r}>0,
    \quad
    \sum_{r=0}^{k-1}\lambda_{\ell,i,r}
    =
    \frac{
        \sum_{r=0}^{k-1}\mathfrak p_{\ell,i}^{\,r}
    }{
        \zeta_{\ell,i}
    }
    =
    1.
\]
Set
\[
    \xi_{\ell,i,0}:=0,
    \quad
    \xi_{\ell,i,r}
    :=
    \sum_{u=0}^{r-1}\lambda_{\ell,i,u},
    \quad r\in[k].
\]
Thus
\[
    0=\xi_{\ell,i,0}
    <
    \xi_{\ell,i,1}
    <
    \cdots
    <
    \xi_{\ell,i,k}=1.
\]

For $r\in\{0,\ldots,k-1\}$, define
\[
    \gamma_{\ell,i,r}(t)
    :=
    \frac{1-\varepsilon_r}{2}
    +
    \frac{\varepsilon_r}{\lambda_{\ell,i,r}}
    (t-\xi_{\ell,i,r}).
\]
Let
\[
    g_{\ell,i}(t):=\gamma_{\ell,i,r}(t),
    \quad
    t\in[\xi_{\ell,i,r},\xi_{\ell,i,r+1}],
\]
and extend the first and last affine formulas to
$(-\infty,0]$ and $[1,\infty)$, respectively.

For $r\in[K]$,
\[
\begin{aligned}
    \gamma_{\ell,i,r-1}(\xi_{\ell,i,r})
    &=
    \frac{1-\varepsilon_{r-1}}2
    +
    \frac{\varepsilon_{r-1}}{\lambda_{\ell,i,r-1}}
    \bigl(
        \xi_{\ell,i,r}-\xi_{\ell,i,r-1}
    \bigr)\\
    &=
    \frac{1-\varepsilon_{r-1}}2+\varepsilon_{r-1}\\
    &=
    \frac{1+\varepsilon_{r-1}}2\\
    &=
    \frac{1-\varepsilon_r}2\\
    &=
    \gamma_{\ell,i,r}(\xi_{\ell,i,r}).
\end{aligned}
\]
Hence $g_{\ell,i}$ is continuous. Moreover,
\[
    \gamma'_{\ell,i,r}
    =
    \frac{\varepsilon_r}{\lambda_{\ell,i,r}}
    =
    \varepsilon_r\zeta_{\ell,i}
    \mathfrak p_{\ell,i}^{-r},
\]
and
\[
\begin{aligned}
    &g'_{\ell,i,+}(\xi_{\ell,i,r})
    -
    g'_{\ell,i,-}(\xi_{\ell,i,r})\\
    &\quad=
    \frac{\varepsilon_r}{\lambda_{\ell,i,r}}
    -
    \frac{\varepsilon_{r-1}}{\lambda_{\ell,i,r-1}}\\
    &\quad=
    \varepsilon_r
    \left(
        \frac1{\lambda_{\ell,i,r}}
        +
        \frac1{\lambda_{\ell,i,r-1}}
    \right)\\
    &\quad\neq0.
\end{aligned}
\]
The affine extensions introduce no additional knots, so
\[
    \mathcal K(g_{\ell,i})
    =
    \{\xi_{\ell,i,1},\ldots,\xi_{\ell,i,K}\},
    \quad
    |\mathcal K(g_{\ell,i})|=K.
\]

Define
\[
    I_{\ell,i,r}
    :=
    (\xi_{\ell,i,r},\xi_{\ell,i,r+1}).
\]
The endpoint values satisfy
\[
\begin{aligned}
    \gamma_{\ell,i,r}(\xi_{\ell,i,r})
    &=
    \frac{1-\varepsilon_r}{2},\\
    \gamma_{\ell,i,r}(\xi_{\ell,i,r+1})
    &=
    \frac{1-\varepsilon_r}{2}+\varepsilon_r
    =
    \frac{1+\varepsilon_r}{2}.
\end{aligned}
\]
Thus
\[
    g_{\ell,i}|_{I_{\ell,i,r}}:
    I_{\ell,i,r}\longrightarrow(0,1)
\]
is an affine bijection.

For $\ell\in[T]$, $j\in[n_\ell]$, and
$i\in[n_{\ell-1}]$, set
\[
    \phi_{\ell,j,i}(t)
    :=
    \begin{cases}
        |G_{\ell,j}|^{-1}g_{\ell,i}(t),
        &j\leq\rho_\ell,\ i\in G_{\ell,j},\\
        0,&\text{otherwise}.
    \end{cases}
\]
Then
\[
    z_j^{(\ell)}(x)
    =
    \begin{cases}
        \displaystyle
        \frac1{|G_{\ell,j}|}
        \sum_{i\in G_{\ell,j}}
        g_{\ell,i}\!\left(z_i^{(\ell-1)}(x)\right),
        &j\leq\rho_\ell,\\[0.8em]
        0,&j>\rho_\ell.
    \end{cases}
\]
Choose
\[
    \mathcal S_{\ell,i}
    :=
    \begin{cases}
        \mathcal K(g_{\ell,i}),&i\leq\rho_{\ell-1},\\
        \varnothing,&i>\rho_{\ell-1}.
    \end{cases}
\]
Since multiplication by $|G_{\ell,j}|^{-1}>0$
preserves the knot set,
\[
    \mathcal K(\phi_{\ell,j,i})
    \subseteq\mathcal S_{\ell,i},
    \quad
    |\mathcal S_{\ell,i}|\leq K.
\]
Therefore
\[
    f\in\mathcal F_{\mathrm{sh}}(\mathbf n,k).
\]

Define
\[
    \widehat F_\ell(x)
    :=
    \bigl(
        z_1^{(\ell)}(x),\ldots,
        z_{\rho_\ell}^{(\ell)}(x)
    \bigr),
    \quad \ell\in\{0,\ldots,T\}.
\]
Identifying $\R^1$ with $\R$,
\[
    \widehat F_0(x)=x,
    \quad
    \widehat F_T=f.
\]

Set
\[
    \mathscr P_0:=\{(0,1)^d\}.
\]
We construct pairwise disjoint nonempty open convex
polyhedral cells satisfying
\[
    F_\ell|_P\ \text{is affine},
    \quad
    \widehat F_\ell(P)=(0,1)^{\rho_\ell},
    \quad P\in\mathscr P_\ell.
\]
These properties hold for $\ell=0$.

Suppose they hold for $\ell-1$.
For $P\in\mathscr P_{\ell-1}$, write
\[
    F_{\ell-1}(x)=W_Px+c_P,
    \quad x\in P,
\]
where
$W_P\in\R^{n_{\ell-1}\times d}$ and
$c_P\in\R^{n_{\ell-1}}$.
For
\[
    \bm a=(a_1,\ldots,a_{\rho_{\ell-1}})
    \in\{0,\ldots,k-1\}^{\rho_{\ell-1}},
\]
set
\[
    B_{\ell,\bm a}
    :=
    \prod_{i=1}^{\rho_{\ell-1}}I_{\ell,i,a_i},
    \quad
    P_{\bm a}
    :=
    P\cap
    \widehat F_{\ell-1}^{-1}(B_{\ell,\bm a}).
\]
Then
\[
\begin{aligned}
    \widehat F_{\ell-1}(P_{\bm a})
    &=
    \widehat F_{\ell-1}(P)
    \cap B_{\ell,\bm a}\\
    &=
    (0,1)^{\rho_{\ell-1}}
    \cap B_{\ell,\bm a}\\
    &=
    B_{\ell,\bm a}\\
    &\neq\varnothing.
\end{aligned}
\]
Consequently $P_{\bm a}\neq\varnothing$. Its explicit form is
\[
    P_{\bm a}
    =
    P\cap
    \bigcap_{i=1}^{\rho_{\ell-1}}
    \left\{
        x:
        \xi_{\ell,i,a_i}
        <
        (W_P)_{i,:}x+(c_P)_i
        <
        \xi_{\ell,i,a_i+1}
    \right\}.
\]
It is therefore open, convex, polyhedral, and full-dimensional.

For $x\in P_{\bm a}$ and $j\in[\rho_\ell]$,
\[
\begin{aligned}
    z_j^{(\ell)}(x)
    &=
    \frac1{|G_{\ell,j}|}
    \sum_{i\in G_{\ell,j}}
    \gamma_{\ell,i,a_i}
    \!\left((W_P)_{i,:}x+(c_P)_i\right)\\
    &=
    \frac1{|G_{\ell,j}|}
    \sum_{i\in G_{\ell,j}}
    \left[
        \frac{1-\varepsilon_{a_i}}2
        +
        \frac{\varepsilon_{a_i}}{\lambda_{\ell,i,a_i}}
        \left(
            (W_P)_{i,:}x+(c_P)_i-\xi_{\ell,i,a_i}
        \right)
    \right].
\end{aligned}
\]
Together with $z_j^{(\ell)}=0$ for $j>\rho_\ell$,
this proves that $F_\ell|_{P_{\bm a}}$ is affine.

Since every folded coordinate belongs to $(0,1)$,
\[
    \widehat F_\ell(P_{\bm a})
    \subseteq(0,1)^{\rho_\ell}.
\]
For the reverse inclusion, fix $y\in(0,1)^{\rho_\ell}$.
For $i\in[\rho_{\ell-1}]$, set
\[
    u_i:=y_{\pi_\ell(i)},
    \quad
    t_i
    :=
    \left(
        g_{\ell,i}|_{I_{\ell,i,a_i}}
    \right)^{-1}(u_i).
\]
Then
\[
    (t_1,\ldots,t_{\rho_{\ell-1}})
    \in B_{\ell,\bm a}
    =
    \widehat F_{\ell-1}(P_{\bm a}).
\]
Hence some $x\in P_{\bm a}$ satisfies
$z_i^{(\ell-1)}(x)=t_i$ for every active coordinate.
For $j\in[\rho_\ell]$,
\[
\begin{aligned}
    z_j^{(\ell)}(x)
    &=
    \frac1{|G_{\ell,j}|}
    \sum_{i\in G_{\ell,j}}g_{\ell,i}(t_i)\\
    &=
    \frac1{|G_{\ell,j}|}
    \sum_{i\in G_{\ell,j}}u_i\\
    &=
    \frac1{|G_{\ell,j}|}
    \sum_{i\in G_{\ell,j}}y_j\\
    &=y_j.
\end{aligned}
\]
Thus
\[
    \widehat F_\ell(P_{\bm a})
    =
    (0,1)^{\rho_\ell}.
\]

Distinct branch tuples give disjoint children.
Children of distinct parents are also disjoint.
Define
\[
    \mathscr P_\ell
    :=
    \left\{
        P_{\bm a}:
        P\in\mathscr P_{\ell-1},\
        \bm a\in\{0,\ldots,k-1\}^{\rho_{\ell-1}}
    \right\}.
\]
Since all children are nonempty,
\[
\begin{aligned}
    |\mathscr P_\ell|
    &=
    \sum_{P\in\mathscr P_{\ell-1}}
    \left|
        \{0,\ldots,k-1\}^{\rho_{\ell-1}}
    \right|\\
    &=
    |\mathscr P_{\ell-1}|k^{\rho_{\ell-1}}\\
    &=
    \prod_{s=1}^{\ell}k^{\rho_{s-1}}\\
    &=
    k^{\sum_{s=1}^{\ell}\rho_{s-1}}.
\end{aligned}
\]

Label a final cell by its selected branch indices
\[
    \alpha=(\alpha_{\ell,i})_{
        \ell\in[T],\,i\in[\rho_{\ell-1}]
    }.
\]
The recursive construction gives
\[
    P_\alpha
    =
    (0,1)^d
    \cap
    \bigcap_{\ell=1}^{T}
    \bigcap_{i=1}^{\rho_{\ell-1}}
    \left\{
        x:
        z_i^{(\ell-1)}(x)
        \in I_{\ell,i,\alpha_{\ell,i}}
    \right\}.
\]
Thus distinct cells have distinct arrays $\alpha$.

For $h\in[d]$, define
\[
    i_0(h):=h,
    \quad
    i_\ell(h):=\pi_\ell(i_{\ell-1}(h)),
    \quad \ell\in[T].
\]
Induction gives
\[
\begin{aligned}
    \{i_0(h):h\in[d]\}
    &=[\rho_0],\\
    \{i_\ell(h):h\in[d]\}
    &=
    \pi_\ell\bigl(
        \{i_{\ell-1}(h):h\in[d]\}
    \bigr)\\
    &=
    \pi_\ell([\rho_{\ell-1}])\\
    &=
    [\rho_\ell].
\end{aligned}
\]

On $P_\alpha$, for $j\in[\rho_\ell]$,
\[
\begin{aligned}
    \partial_{x_h}z_j^{(\ell)}
    &=
    \frac1{|G_{\ell,j}|}
    \sum_{i\in G_{\ell,j}}
    g'_{\ell,i}(z_i^{(\ell-1)})
    \partial_{x_h}z_i^{(\ell-1)}\\
    &=
    \frac1{|G_{\ell,j}|}
    \sum_{i\in G_{\ell,j}}
    \frac{
        \varepsilon_{\alpha_{\ell,i}}\zeta_{\ell,i}
    }{
        \mathfrak p_{\ell,i}^{\alpha_{\ell,i}}
    }
    \partial_{x_h}z_i^{(\ell-1)}.
\end{aligned}
\]
At layer zero,
\[
    \partial_{x_h}z_h^{(0)}=1,
    \quad
    \partial_{x_h}z_i^{(0)}=0
    \quad(i\neq h).
\]
Suppose only the coordinate $i_{\ell-1}(h)$ has a nonzero
derivative with respect to $x_h$ at layer $\ell-1$.
For $j\leq\rho_\ell$,
\[
    j\neq i_\ell(h)
    \quad\Longrightarrow\quad
    i_{\ell-1}(h)\notin G_{\ell,j}
    \quad\Longrightarrow\quad
    \partial_{x_h}z_j^{(\ell)}=0.
\]
Coordinates $j>\rho_\ell$ are identically zero. For the
remaining coordinate,
\[
\begin{aligned}
    \partial_{x_h}z_{i_\ell(h)}^{(\ell)}
    &=
    \frac1{|G_{\ell,i_\ell(h)}|}
    \sum_{i\in G_{\ell,i_\ell(h)}}
    \frac{
        \varepsilon_{\alpha_{\ell,i}}\zeta_{\ell,i}
    }{
        \mathfrak p_{\ell,i}^{\alpha_{\ell,i}}
    }
    \partial_{x_h}z_i^{(\ell-1)}\\
    &=
    \frac{
        \varepsilon_{\alpha_{\ell,i_{\ell-1}(h)}}
        \zeta_{\ell,i_{\ell-1}(h)}
    }{
        |G_{\ell,i_\ell(h)}|\,
        \mathfrak p_{\ell,i_{\ell-1}(h)}^{
            \alpha_{\ell,i_{\ell-1}(h)}
        }
    }
    \partial_{x_h}z_{i_{\ell-1}(h)}^{(\ell-1)}\\
    &\neq0.
\end{aligned}
\]
The derivative property therefore holds at every layer.
Since $i_T(h)=1$ and $f=z_1^{(T)}$,
\[
\begin{aligned}
    \left|
        \left.\partial_{x_h}f\right|_{P_\alpha}
    \right|
    &=
    \prod_{\ell=1}^{T}
    \left|
        \frac{
            \varepsilon_{\alpha_{\ell,i_{\ell-1}(h)}}
            \zeta_{\ell,i_{\ell-1}(h)}
        }{
            |G_{\ell,i_\ell(h)}|\,
            \mathfrak p_{\ell,i_{\ell-1}(h)}^{
                \alpha_{\ell,i_{\ell-1}(h)}
            }
        }
    \right|\\
    &=
    \prod_{\ell=1}^{T}
    \frac{
        \zeta_{\ell,i_{\ell-1}(h)}
    }{
        |G_{\ell,i_\ell(h)}|\,
        \mathfrak p_{\ell,i_{\ell-1}(h)}^{
            \alpha_{\ell,i_{\ell-1}(h)}
        }
    }\\
    &>0.
\end{aligned}
\]

Let $P_\alpha\neq P_\beta$ and choose $(\ell_*,i_*)$
with
\[
    \alpha_{\ell_*,i_*}\neq\beta_{\ell_*,i_*}.
\]
The surjectivity of $h\mapsto i_{\ell_*-1}(h)$ gives
$h\in[d]$ such that
\[
    i_{\ell_*-1}(h)=i_*.
\]
If the two final gradients were equal, then
\[
\begin{aligned}
    1
    &=
    \frac{
        \left|
            \left.\partial_{x_h}f\right|_{P_\alpha}
        \right|
    }{
        \left|
            \left.\partial_{x_h}f\right|_{P_\beta}
        \right|
    }\\
    &=
    \prod_{\ell=1}^{T}
    \mathfrak p_{\ell,i_{\ell-1}(h)}^{
        \beta_{\ell,i_{\ell-1}(h)}
        -
        \alpha_{\ell,i_{\ell-1}(h)}
    }.
\end{aligned}
\]
Hence
\[
    \prod_{\ell=1}^{T}
    \mathfrak p_{\ell,i_{\ell-1}(h)}^{
        \alpha_{\ell,i_{\ell-1}(h)}
    }
    =
    \prod_{\ell=1}^{T}
    \mathfrak p_{\ell,i_{\ell-1}(h)}^{
        \beta_{\ell,i_{\ell-1}(h)}
    }.
\]
All primes appearing in these products are distinct.
Uniqueness of prime factorization implies
\[
    \alpha_{\ell,i_{\ell-1}(h)}
    =
    \beta_{\ell,i_{\ell-1}(h)},
    \quad \ell\in[T].
\]
In particular,
\[
    \alpha_{\ell_*,i_*}
    =
    \beta_{\ell_*,i_*},
\]
a contradiction. Therefore
\[
    \left.\nabla f\right|_{P_\alpha}
    \neq
    \left.\nabla f\right|_{P_\beta}
    \quad(\alpha\neq\beta).
\]

For each $P_\alpha$, let $L_\alpha:\R^d\to\R$
be the affine extension of $f|_{P_\alpha}$ and set
\[
    \Omega_\alpha
    :=
    \operatorname{int}
    \{x\in\R^d:f(x)=L_\alpha(x)\}.
\]
Since $P_\alpha$ is open and connected,
it lies in one connected component $U_\alpha$ of
$\Omega_\alpha$. This component is open.

Let $V$ be a connected open affine set containing
$U_\alpha$, with affine formula $L_V$. Then
\[
\begin{aligned}
    P_\alpha\subseteq U_\alpha\subseteq V
    &\quad\Longrightarrow\quad
    L_V=L_\alpha\ \text{on }P_\alpha\\
    &\quad\Longrightarrow\quad
    L_V=L_\alpha\ \text{on }\R^d\\
    &\quad\Longrightarrow\quad
    V\subseteq\Omega_\alpha\\
    &\quad\Longrightarrow\quad
    V\subseteq U_\alpha.
\end{aligned}
\]
Thus $V=U_\alpha$, and $U_\alpha$ is a linear region.
Different cells have different gradients, so
\[
    \alpha\neq\beta
    \quad\Longrightarrow\quad
    U_\alpha\neq U_\beta.
\]
Consequently,
\[
\begin{aligned}
    \mathcal R(f)
    &\geq|\mathscr P_T|\\
    &=
    \prod_{\ell=1}^{T}k^{\rho_{\ell-1}}\\
    &=
    k^{\sum_{\ell=1}^{T}\rho_{\ell-1}}.
\end{aligned}
\]
Using
$f\in\mathcal F_{\mathrm{sh}}(\mathbf n,k)
\subseteq\mathcal F_{\mathrm{ind}}(\mathbf n,k)$,
we obtain
\[
    \mathcal R^\star_{\mathrm{ind}}(\mathbf n,k)
    \geq
    \mathcal R^\star_{\mathrm{sh}}(\mathbf n,k)
    \geq
    \mathcal R(f)
    \geq
    k^{\sum_{\ell=1}^{T}\rho_{\ell-1}}.
\]
This completes the proof.
\end{proof}

\subsection{Proof of Corollary~\ref{cor:tight-kan-bounds}}

\begin{proof}
By Theorems~\ref{thm:global-upper}
and~\ref{thm:global-lower}, for every $k\geq2$,
\[
\begin{aligned}
    1
    &\leq
    \frac{
        \mathcal R^\star_{\mathrm{sh}}(\mathbf n,k)
    }{
        k^{\sum_{\ell=1}^{T}\rho_{\ell-1}}
    }\\
    &\leq
    \frac{
        \mathcal R^\star_{\mathrm{ind}}(\mathbf n,k)
    }{
        k^{\sum_{\ell=1}^{T}\rho_{\ell-1}}
    }\\
    &\leq
    \prod_{\ell=1}^{T}
    \left[
        \sum_{s=0}^{\rho_{\ell-1}}
        \binom{n_{\ell-1}}s n_\ell^s
    \right]\\
    &<\infty.
\end{aligned}
\]
For fixed $\mathbf n$, the final product is independent
of $k$. Therefore, as $k\to\infty$,
\[
\begin{aligned}
    \mathcal R^\star_{\mathrm{sh}}(\mathbf n,k)
    &=
    \Theta_{\mathbf n}\!\left(
        k^{\sum_{\ell=1}^{T}\rho_{\ell-1}}
    \right),\\
    \mathcal R^\star_{\mathrm{ind}}(\mathbf n,k)
    &=
    \Theta_{\mathbf n}\!\left(
        k^{\sum_{\ell=1}^{T}\rho_{\ell-1}}
    \right).
\end{aligned}
\]
This completes the proof.
\end{proof}

\subsection{Proof of Corollary~\ref{cor:aistats-comparison}}

\begin{proof}
For every $\ell\in[T]$,
\[
    n_\ell\geq1,
    \quad
    \rho_{\ell-1}\leq n_{\ell-1},
    \quad
    K=k-1\geq1.
\]
For every integer $s$ with $0\leq s\leq\rho_{\ell-1}$,
\[
    0\leq K^s\leq(n_\ell K)^s.
\]
Moreover,
\[
\begin{aligned}
    k^{n_\ell}
    &=(1+K)^{n_\ell}\\
    &=\sum_{u=0}^{n_\ell}\binom{n_\ell}uK^u\\
    &=1+n_\ell K+
      \sum_{u=2}^{n_\ell}\binom{n_\ell}uK^u\\
    &\geq1+n_\ell K.
\end{aligned}
\]
Consequently,
\[
\begin{aligned}
    &\sum_{s=0}^{\rho_{\ell-1}}
    \binom{n_{\ell-1}}sK^s\\
    &\quad\leq
    \sum_{s=0}^{\rho_{\ell-1}}
    \binom{n_{\ell-1}}s(n_\ell K)^s\\
    &\quad\leq
    \sum_{s=0}^{n_{\ell-1}}
    \binom{n_{\ell-1}}s(n_\ell K)^s\\
    &\quad=
    (1+n_\ell K)^{n_{\ell-1}}\\
    &\quad\leq
    \bigl(k^{n_\ell}\bigr)^{n_{\ell-1}}\\
    &\quad=
    k^{n_{\ell-1}n_\ell}.
\end{aligned}
\]
Multiplying over $\ell\in[T]$ gives
\[
\begin{aligned}
    &\prod_{\ell=1}^{T}
    \left[
        \sum_{s=0}^{\rho_{\ell-1}}
        \binom{n_{\ell-1}}sK^s
    \right]\\
    &\quad\leq
    \prod_{\ell=1}^{T}
    \left[
        \sum_{s=0}^{\rho_{\ell-1}}
        \binom{n_{\ell-1}}s(n_\ell K)^s
    \right]\\
    &\quad\leq
    \prod_{\ell=1}^{T}k^{n_{\ell-1}n_\ell}\\
    &\quad=
    k^{\sum_{\ell=1}^{T}n_{\ell-1}n_\ell}.
\end{aligned}
\]
Substituting $K=k-1$ proves the finite inequalities.

Fix $\mathbf n$ and
$\diamond\in\{\mathrm{ind},\mathrm{sh}\}$. Define
\[
    C_{\diamond,\mathbf n}
    :=
    \begin{cases}
        \displaystyle
        \prod_{\ell=1}^{T}
        \left[
            \sum_{s=0}^{\rho_{\ell-1}}
            \binom{n_{\ell-1}}s n_\ell^s
        \right],
        &\diamond=\mathrm{ind},\\[1em]
        \displaystyle
        \prod_{\ell=1}^{T}
        \left[
            \sum_{s=0}^{\rho_{\ell-1}}
            \binom{n_{\ell-1}}s
        \right],
        &\diamond=\mathrm{sh}.
    \end{cases}
\]
Every factor contains the term $s=0$, so
\[
    1\leq C_{\diamond,\mathbf n}<\infty.
\]
This constant is independent of $k$. By
Theorems~\ref{thm:global-upper}
and~\ref{thm:global-lower},
\[
    k^{\sum_{\ell=1}^{T}\rho_{\ell-1}}
    \leq
    \mathcal R^\star_\diamond(\mathbf n,k)
    \leq
    C_{\diamond,\mathbf n}
    k^{\sum_{\ell=1}^{T}\rho_{\ell-1}}.
\]
Dividing by
$k^{\sum_{\ell=1}^{T}n_{\ell-1}n_\ell}>0$ yields
\[
\begin{aligned}
    k^{
        \sum_{\ell=1}^{T}\rho_{\ell-1}
        -
        \sum_{\ell=1}^{T}n_{\ell-1}n_\ell
    }
    &\leq
    \frac{
        \mathcal R^\star_\diamond(\mathbf n,k)
    }{
        k^{\sum_{\ell=1}^{T}n_{\ell-1}n_\ell}
    }\\
    &\leq
    C_{\diamond,\mathbf n}
    k^{
        \sum_{\ell=1}^{T}\rho_{\ell-1}
        -
        \sum_{\ell=1}^{T}n_{\ell-1}n_\ell
    }.
\end{aligned}
\]
Therefore, as $k\to\infty$,
\[
    \frac{
        \mathcal R^\star_\diamond(\mathbf n,k)
    }{
        k^{\sum_{\ell=1}^{T}n_{\ell-1}n_\ell}
    }
    =
    \Theta_{\mathbf n}\!\left(
        k^{
            \sum_{\ell=1}^{T}\rho_{\ell-1}
            -
            \sum_{\ell=1}^{T}n_{\ell-1}n_\ell
        }
    \right).
\]

The exponent difference satisfies
\[
\begin{aligned}
    &\sum_{\ell=1}^{T}n_{\ell-1}n_\ell
    -
    \sum_{\ell=1}^{T}\rho_{\ell-1}\\
    &\quad=
    \sum_{\ell=1}^{T}
    \bigl(n_{\ell-1}n_\ell-\rho_{\ell-1}\bigr)\\
    &\quad=
    \sum_{\ell=1}^{T}
    \left[
        n_{\ell-1}(n_\ell-1)
        +
        n_{\ell-1}-\rho_{\ell-1}
    \right]\\
    &\quad\geq0.
\end{aligned}
\]
If some hidden layer $\ell_*\in[T-1]$ satisfies
$n_{\ell_*}\geq2$, then
\[
\begin{aligned}
    &\sum_{\ell=1}^{T}n_{\ell-1}n_\ell
    -
    \sum_{\ell=1}^{T}\rho_{\ell-1}\\
    &\quad\geq
    n_{\ell_*-1}(n_{\ell_*}-1)\\
    &\quad\geq n_{\ell_*-1}\\
    &\quad\geq1.
\end{aligned}
\]
Hence
\[
\begin{aligned}
    0
    &\leq
    \frac{
        \mathcal R^\star_\diamond(\mathbf n,k)
    }{
        k^{\sum_{\ell=1}^{T}n_{\ell-1}n_\ell}
    }\\
    &\leq
    C_{\diamond,\mathbf n}
    k^{
        \sum_{\ell=1}^{T}\rho_{\ell-1}
        -
        \sum_{\ell=1}^{T}n_{\ell-1}n_\ell
    }\\
    &\leq
    \frac{C_{\diamond,\mathbf n}}{k}
    \longrightarrow0.
\end{aligned}
\]

If $n_\ell=1$ for every $\ell\in[T]$, then
\[
    \rho_0=d,
    \quad
    \rho_{\ell-1}=1
    \quad\text{for }2\leq\ell\leq T,
\]
and
\[
\begin{aligned}
    \sum_{\ell=1}^{T}\rho_{\ell-1}
    &=d+\sum_{\ell=2}^{T}1\\
    &=d+T-1,\\
    \sum_{\ell=1}^{T}n_{\ell-1}n_\ell
    &=d\cdot1+\sum_{\ell=2}^{T}1\cdot1\\
    &=d+T-1.
\end{aligned}
\]
The upper and lower bounds give
\[
\begin{aligned}
    k^{d+T-1}
    &\leq
    \mathcal R^\star_{\mathrm{sh}}(\mathbf n,k)\\
    &\leq
    \mathcal R^\star_{\mathrm{ind}}(\mathbf n,k)\\
    &\leq
    \left[
        \sum_{s=0}^{d}\binom dsK^s
    \right]
    \prod_{\ell=2}^{T}
    \left[
        \sum_{s=0}^{1}\binom1sK^s
    \right]\\
    &=(1+K)^d
      \prod_{\ell=2}^{T}(1+K)\\
    &=k^d k^{T-1}\\
    &=k^{d+T-1}.
\end{aligned}
\]
Thus both maximum region counts equal the bound of
Schoots et al. In particular, $T=1$ gives $k^d$.

Finally, suppose $T\geq2$ and
$n_1=\cdots=n_{T-1}=n$. Then
\[
    \rho_0=d,
    \quad
    \rho_{\ell-1}=\min\{d,n\}
    \quad\text{for }2\leq\ell\leq T.
\]
Therefore,
\[
\begin{aligned}
    \sum_{\ell=1}^{T}\rho_{\ell-1}
    &=d+\sum_{\ell=2}^{T}\min\{d,n\}\\
    &=d+(T-1)\min\{d,n\},
\end{aligned}
\]
and
\[
\begin{aligned}
    \sum_{\ell=1}^{T}n_{\ell-1}n_\ell
    &=n_0n_1+
      \sum_{\ell=2}^{T-1}n_{\ell-1}n_\ell+
      n_{T-1}n_T\\
    &=dn+\sum_{\ell=2}^{T-1}n^2+n\\
    &=dn+(T-2)n^2+n.
\end{aligned}
\]
Substitution into the ratio gives
\[
    \frac{
        \mathcal R^\star_\diamond(\mathbf n,k)
    }{
        k^{dn+(T-2)n^2+n}
    }
    =
    \Theta_{d,n,T}\!\left(
        k^{
            d+(T-1)\min\{d,n\}
            -dn-(T-2)n^2-n
        }
    \right).
\]
If $n\geq d$, then
\[
    d+(T-1)\min\{d,n\}
    =
    d+(T-1)d
    =
    dT.
\]
This completes the proof.
\end{proof}

\subsection{Proof of Proposition~\ref{prop:kan-relu-equivalence}}

\begin{proof}
Fix one edge function $\phi_{j,i}$ and let
$s_{j,i}:=|\mathcal K(\phi_{j,i})|$. Write its knots as
\[
    \tau_{j,i,1}<\cdots<\tau_{j,i,s_{j,i}},
    \quad s_{j,i}\leq K,
\]
with
\[
    \tau_{j,i,0}:=-\infty,
    \quad
    \tau_{j,i,s_{j,i}+1}:=+\infty.
\]
Let $b_{j,i,u}$ be its slope on
\[
    (\tau_{j,i,u},\tau_{j,i,u+1}),
    \quad u\in\{0,\ldots,s_{j,i}\}.
\]
This includes $s_{j,i}=0$, when the sole interval is $\R$.

Choose
$t_0\in(\tau_{j,i,0},\tau_{j,i,1})$ and define
\[
    a_{j,i}
    :=
    \phi_{j,i}(t_0)-b_{j,i,0}t_0,
\]
where the first interval is understood as $\R$ if
$s_{j,i}=0$. For $r\in[s_{j,i}]$, let
\[
    c_{j,i,r}
    :=
    b_{j,i,r}-b_{j,i,r-1}.
\]
Define
\[
    \psi_{j,i}(t)
    :=
    a_{j,i}+b_{j,i,0}t+
    \sum_{r=1}^{s_{j,i}}
    c_{j,i,r}\sigma(t-\tau_{j,i,r}).
\]
For $t\in(\tau_{j,i,u},\tau_{j,i,u+1})$,
\[
\begin{aligned}
    \psi_{j,i}'(t)
    &=
    b_{j,i,0}+\sum_{r=1}^{u}c_{j,i,r}\\
    &=
    b_{j,i,0}+
    \sum_{r=1}^{u}
    (b_{j,i,r}-b_{j,i,r-1})\\
    &=
    b_{j,i,u}\\
    &=
    \phi_{j,i}'(t).
\end{aligned}
\]
Thus $\psi_{j,i}-\phi_{j,i}$ is constant on each knot interval.
Both functions are continuous, so these constants agree across
all knots. At $t_0$,
\[
\begin{aligned}
    \psi_{j,i}(t_0)
    &=
    a_{j,i}+b_{j,i,0}t_0\\
    &=
    \phi_{j,i}(t_0).
\end{aligned}
\]
Therefore
\[
    \phi_{j,i}(t)
    =
    a_{j,i}+b_{j,i,0}t+
    \sum_{r=1}^{s_{j,i}}
    c_{j,i,r}\sigma(t-\tau_{j,i,r})
\]
for every $t\in\R$.

Using
\[
    t=\sigma(t)-\sigma(-t),
\]
the KAN output satisfies
\[
\begin{aligned}
    [\Phi(z)]_j
    &=
    \sum_{i=1}^{p}\phi_{j,i}(z_i)\\
    &=
    \sum_{i=1}^{p}a_{j,i}
    +
    \sum_{i=1}^{p}b_{j,i,0}\sigma(z_i)
    -
    \sum_{i=1}^{p}b_{j,i,0}\sigma(-z_i)\\
    &\quad+
    \sum_{i=1}^{p}
    \sum_{r=1}^{s_{j,i}}
    c_{j,i,r}\sigma(z_i-\tau_{j,i,r}).
\end{aligned}
\]
The $2p$ features
\[
    \sigma(z_i),\ \sigma(-z_i),
    \quad i\in[p],
\]
are shared across all outputs. The remaining features number
at most
\[
\begin{aligned}
    \sum_{j=1}^{q}\sum_{i=1}^{p}s_{j,i}
    &\leq
    \sum_{j=1}^{q}\sum_{i=1}^{p}K\\
    &=pqK.
\end{aligned}
\]
An affine output layer combines these features, so the required
hidden width is at most
\[
    pqK+2p.
\]

In the source-shared-knot model, write
\[
    \phi_{j,i}(t)
    =
    a_{j,i}+b_{j,i,0}t+
    \sum_{\tau\in\mathcal S_i}
    c_{j,i,\tau}\sigma(t-\tau),
\]
where $c_{j,i,\tau}=0$ for candidate knots not used by the edge.
For each source $i$, the features
$\sigma(z_i-\tau)$, $\tau\in\mathcal S_i$, are shared across
all outputs. Hence the width is at most
\[
\begin{aligned}
    2p+\sum_{i=1}^{p}|\mathcal S_i|
    &\leq2p+\sum_{i=1}^{p}K\\
    &=pK+2p.
\end{aligned}
\]

Conversely, consider a ReLU layer
\[
    y_j
    =
    \sigma\left(
        \sum_{i=1}^{p}w_{j,i}z_i+b_j
    \right),
    \quad j\in[q].
\]
Define $\Psi:\R^p\to\R^q$ using the affine edge functions
\[
    \psi_{j,i}(t)
    :=
    w_{j,i}t+\frac{b_j}{p}.
\]
Then
\[
\begin{aligned}
    [\Psi(z)]_j
    &=
    \sum_{i=1}^{p}
    \left(w_{j,i}z_i+\frac{b_j}{p}\right)\\
    &=
    \sum_{i=1}^{p}w_{j,i}z_i+
    p\frac{b_j}{p}\\
    &=
    \sum_{i=1}^{p}w_{j,i}z_i+b_j.
\end{aligned}
\]
Define $\Gamma:\R^q\to\R^q$ by
\[
    \gamma_{j,i}(t)
    :=
    \begin{cases}
        \sigma(t),&i=j,\\
        0,&i\neq j.
    \end{cases}
\]
It follows that
\[
\begin{aligned}
    [\Gamma(\Psi(z))]_j
    &=
    \sum_{i=1}^{q}\gamma_{j,i}([\Psi(z)]_i)\\
    &=
    \sigma([\Psi(z)]_j)\\
    &=
    \sigma\left(
        \sum_{i=1}^{p}w_{j,i}z_i+b_j
    \right)\\
    &=y_j.
\end{aligned}
\]
For $\Psi$, choose empty candidate knot sets. For $\Gamma$,
choose $\{0\}$ at every source. Every edge then has at most
two affine segments and obeys the source-sharing condition.
Thus the conversion uses two source-shared-knot KAN layers
with widths $(p,q,q)$ and at most two affine segments
per edge ($k=2$).
\end{proof}

\subsection{Proof of Theorem~\ref{thm:parameter-budget-comparison}}

\begin{proof}
Fix $d,T\in\mathbb N$, and define
\[
    \chi_d:=\sum_{j=0}^{d}\frac1{j!},
    \quad
    \nu_T:=\left\lceil\frac T2\right\rceil.
\]

Let $f\in\mathcal F_{\mathrm{ind}}(\mathbf n,k)$ satisfy
$n_0=d$, $n_T=1$, and
$\Lambda_{\mathrm{KAN}}(\mathbf n,k)\leq\Lambda$.
For $\ell\in[T]$, set
\[
    e_\ell:=n_{\ell-1}n_\ell.
\]
For $0\leq j\leq\rho_{\ell-1}\leq n_{\ell-1}$,
\[
\begin{aligned}
    \binom{n_{\ell-1}}j
    \bigl(n_\ell(k-1)\bigr)^j
    &=
    \frac{
        \prod_{u=0}^{j-1}(n_{\ell-1}-u)
    }{j!}
    n_\ell^j(k-1)^j\\
    &\leq
    \frac{
        n_{\ell-1}^j n_\ell^j(k-1)^j
    }{j!}\\
    &=
    \frac{\bigl(e_\ell(k-1)\bigr)^j}{j!}\\
    &\leq
    \frac{(ke_\ell)^j}{j!}.
\end{aligned}
\]
Since $\rho_{\ell-1}\leq d$ and $ke_\ell\geq1$,
\[
\begin{aligned}
    &\sum_{j=0}^{\rho_{\ell-1}}
    \binom{n_{\ell-1}}j
    \bigl(n_\ell(k-1)\bigr)^j\\
    &\quad\leq
    \sum_{j=0}^{\rho_{\ell-1}}
    \frac{(ke_\ell)^j}{j!}\\
    &\quad\leq
    \sum_{j=0}^{d}\frac{(ke_\ell)^j}{j!}\\
    &\quad\leq
    \sum_{j=0}^{d}\frac{(ke_\ell)^d}{j!}\\
    &\quad=
    \chi_d(ke_\ell)^d.
\end{aligned}
\]
Theorem~\ref{thm:global-upper} and the arithmetic--geometric
mean inequality give
\[
\begin{aligned}
    \mathcal R(f)
    &\leq
    \prod_{\ell=1}^{T}
    \left[
        \sum_{j=0}^{\rho_{\ell-1}}
        \binom{n_{\ell-1}}j
        \bigl(n_\ell(k-1)\bigr)^j
    \right]\\
    &\leq
    \prod_{\ell=1}^{T}
    \bigl[\chi_d(ke_\ell)^d\bigr]\\
    &=
    \chi_d^T k^{dT}
    \left(\prod_{\ell=1}^{T}e_\ell\right)^d\\
    &\leq
    \chi_d^T k^{dT}
    \left(
        \frac{\sum_{\ell=1}^{T}e_\ell}{T}
    \right)^{dT}\\
    &=
    \chi_d^T
    \left(
        \frac{2k\sum_{\ell=1}^{T}e_\ell}{2T}
    \right)^{dT}\\
    &=
    \chi_d^T
    \left(
        \frac{\Lambda_{\mathrm{KAN}}(\mathbf n,k)}{2T}
    \right)^{dT}\\
    &\leq
    \chi_d^T
    \left(\frac{\Lambda}{2T}\right)^{dT}.
\end{aligned}
\]
Thus,
\[
    \mathcal R^\star_{\mathrm{KAN}}(d,T,\Lambda)
    \leq
    \chi_d^T
    \left(\frac{\Lambda}{2T}\right)^{dT}.
\]

For the KAN lower bound, choose
\[
    n_\ell=d
    \quad\text{for }0\leq\ell<T,
    \quad
    n_T=1.
\]
Then
\[
\begin{aligned}
    e_{\mathrm{base}}
    &:=
    \sum_{\ell=1}^{T}n_{\ell-1}n_\ell\\
    &=
    \sum_{\ell=1}^{T-1}d^2+d\\
    &=
    (T-1)d^2+d.
\end{aligned}
\]
Define
\[
    k_\Lambda:=
    \left\lfloor
        \frac{\Lambda}{2e_{\mathrm{base}}}
    \right\rfloor.
\]
For $\Lambda\geq4e_{\mathrm{base}}$,
\[
    k_\Lambda
    \geq
    \lfloor2\rfloor
    =
    2,
\]
and
\[
\begin{aligned}
    \Lambda_{\mathrm{KAN}}(\mathbf n,k_\Lambda)
    &=
    2e_{\mathrm{base}}k_\Lambda\\
    &\leq
    2e_{\mathrm{base}}
    \frac{\Lambda}{2e_{\mathrm{base}}}\\
    &=
    \Lambda.
\end{aligned}
\]
Moreover,
\[
\begin{aligned}
    k_\Lambda
    &\geq
    \frac{\Lambda}{2e_{\mathrm{base}}}-1\\
    &=
    \frac{\Lambda}{4e_{\mathrm{base}}}
    +
    \left(
        \frac{\Lambda}{4e_{\mathrm{base}}}-1
    \right)\\
    &\geq
    \frac{\Lambda}{4e_{\mathrm{base}}}.
\end{aligned}
\]
For this architecture,
\[
    \rho_{\ell-1}=d,
    \quad \ell\in[T].
\]
Theorem~\ref{thm:global-lower} therefore yields
\[
\begin{aligned}
    \mathcal R^\star_{\mathrm{KAN}}(d,T,\Lambda)
    &\geq
    \mathcal R^\star_{\mathrm{ind}}(\mathbf n,k_\Lambda)\\
    &\geq
    \mathcal R^\star_{\mathrm{sh}}(\mathbf n,k_\Lambda)\\
    &\geq
    k_\Lambda^{dT}\\
    &\geq
    \left(
        \frac{\Lambda}{4e_{\mathrm{base}}}
    \right)^{dT}.
\end{aligned}
\]
Consequently,
\[
    \mathcal R^\star_{\mathrm{KAN}}(d,T,\Lambda)
    =
    \Theta_{d,T}(\Lambda^{dT}).
\]

Next, let $g:\R^d\to\R$ be a ReLU network with width vector
$\mathbf m=(m_0,\ldots,m_{T+1})$, where
$m_0=d$, $m_{T+1}=1$, and
$\Lambda_{\mathrm{ReLU}}(\mathbf m)\leq\Lambda$.
On each cell produced by the preceding hidden layers, hidden
layer $\ell$ introduces at most $m_\ell$ affine hyperplanes.
The affine output layer introduces no further subdivision.
Hence
\[
    \mathcal R(g)
    \leq
    \prod_{\ell=1}^{T}A_d(m_\ell,1).
\]
For $0\leq j\leq\min\{d,m_\ell\}$,
\[
    \binom{m_\ell}j
    =
    \frac{
        \prod_{u=0}^{j-1}(m_\ell-u)
    }{j!}
    \leq
    \frac{m_\ell^j}{j!}.
\]
Since $m_\ell\geq1$,
\[
\begin{aligned}
    A_d(m_\ell,1)
    &=
    \sum_{j=0}^{\min\{d,m_\ell\}}\binom{m_\ell}j\\
    &\leq
    \sum_{j=0}^{\min\{d,m_\ell\}}
    \frac{m_\ell^j}{j!}\\
    &\leq
    \sum_{j=0}^{d}\frac{m_\ell^j}{j!}\\
    &\leq
    \sum_{j=0}^{d}\frac{m_\ell^d}{j!}\\
    &=
    \chi_d m_\ell^d.
\end{aligned}
\]
Therefore,
\[
\begin{aligned}
    \mathcal R(g)
    &\leq
    \prod_{\ell=1}^{T}\bigl(\chi_d m_\ell^d\bigr)\\
    &=
    \chi_d^T
    \left(\prod_{\ell=1}^{T}m_\ell\right)^d.
\end{aligned}
\]

The parameter constraint gives
\[
\begin{aligned}
    (d+1)m_1
    &\leq
    \Lambda_{\mathrm{ReLU}}(\mathbf m)
    \leq\Lambda,\\
    m_1
    &\leq
    \frac{\Lambda}{d+1}
    \leq\Lambda,
\end{aligned}
\]
and, for $2\leq\ell\leq T$,
\[
\begin{aligned}
    m_{\ell-1}m_\ell
    &\leq
    (m_{\ell-1}+1)m_\ell\\
    &\leq
    \Lambda_{\mathrm{ReLU}}(\mathbf m)\\
    &\leq\Lambda.
\end{aligned}
\]
Thus
\[
\begin{aligned}
    \prod_{\ell=1}^{T}m_\ell
    &=
    \begin{cases}
        \displaystyle
        \prod_{a=1}^{\nu_T}
        (m_{2a-1}m_{2a}),
        &T=2\nu_T,\\[0.8em]
        \displaystyle
        m_1\prod_{a=1}^{\nu_T-1}
        (m_{2a}m_{2a+1}),
        &T=2\nu_T-1,
    \end{cases}\\
    &\leq
    \begin{cases}
        \Lambda^{\nu_T},
        &T=2\nu_T,\\
        \Lambda\Lambda^{\nu_T-1},
        &T=2\nu_T-1,
    \end{cases}\\
    &=
    \Lambda^{\nu_T}.
\end{aligned}
\]
It follows that
\[
    \mathcal R^\star_{\mathrm{ReLU}}(d,T,\Lambda)
    \leq
    \chi_d^T\Lambda^{d\nu_T}.
\]

For the ReLU lower bound, define
\[
    \eta_{d,T}:=(T+1)(d+1),
    \quad
    \mu_\Lambda:=
    \left\lfloor\frac{\Lambda}{\eta_{d,T}}\right\rfloor-1.
\]
For $\Lambda\geq4d\eta_{d,T}$,
\[
\begin{aligned}
    \mu_\Lambda
    &\geq
    \frac{\Lambda}{\eta_{d,T}}-2\\
    &=
    \frac{\Lambda}{2\eta_{d,T}}
    +
    \left(
        \frac{\Lambda}{2\eta_{d,T}}-2
    \right)\\
    &\geq
    \frac{\Lambda}{2\eta_{d,T}}\\
    &\geq2d.
\end{aligned}
\]
Choose $m_0=d$, $m_{T+1}=1$, and
\[
    m_\ell:=
    \begin{cases}
        \mu_\Lambda,&\ell\text{ odd},\\
        d,&\ell\text{ even},
    \end{cases}
    \quad \ell\in[T].
\]
The parameter count of each affine layer is
\[
    (m_{\ell-1}+1)m_\ell
    =
    \begin{cases}
        (d+1)\mu_\Lambda,
        &1\leq\ell\leq T,\ \ell\text{ odd},\\
        d(\mu_\Lambda+1),
        &1\leq\ell\leq T,\ \ell\text{ even},\\
        \mu_\Lambda+1,
        &\ell=T+1,\ T\text{ odd},\\
        d+1,
        &\ell=T+1,\ T\text{ even}.
    \end{cases}
\]
All four expressions are at most
$(d+1)(\mu_\Lambda+1)$. Therefore,
\[
\begin{aligned}
    \Lambda_{\mathrm{ReLU}}(\mathbf m)
    &=
    \sum_{\ell=1}^{T+1}(m_{\ell-1}+1)m_\ell\\
    &\leq
    \sum_{\ell=1}^{T+1}(d+1)(\mu_\Lambda+1)\\
    &=
    (T+1)(d+1)(\mu_\Lambda+1)\\
    &=
    \eta_{d,T}
    \left\lfloor\frac{\Lambda}{\eta_{d,T}}\right\rfloor\\
    &\leq
    \eta_{d,T}\frac{\Lambda}{\eta_{d,T}}\\
    &=
    \Lambda.
\end{aligned}
\]

Since $m_\ell\geq d$ for every $\ell\in[T]$, the construction
underlying
Mont\'ufar et al.~\cite[Theorem~5]{montufar2014number}
provides a final hidden map
$\mathbf h:\R^d\to\R^{m_T}$ satisfying
\[
    \mathcal R(\mathbf h)
    \geq
    \left[
        \prod_{\ell=1}^{T-1}
        \left\lfloor\frac{m_\ell}{d}\right\rfloor^d
    \right]
    A_d(m_T,1).
\]

Choose a finite polyhedral partition $\mathcal P$ on which
$\mathbf h$ is affine. Let $\iota$ be the number of distinct
affine formulas on its full-dimensional cells, and enumerate
their coefficient pairs as
\[
    (D_a,\delta_a),
    \quad
    D_a\in\R^{m_T\times d},
    \quad
    \delta_a\in\R^{m_T},
    \quad a\in[\iota].
\]
For each $P\in\mathcal P$, choose $a(P)\in[\iota]$ such that
\[
    \mathbf h(x)
    =
    D_{a(P)}x+\delta_{a(P)},
    \quad x\in P.
\]

For $1\leq a<b\leq\iota$, distinctness gives
\[
    \operatorname{rank}
    \begin{bmatrix}
        D_a-D_b & \delta_a-\delta_b
    \end{bmatrix}
    \geq1.
\]
Consequently,
\[
\begin{aligned}
    &\dim\left\{
        \omega'\in\R^{m_T}:
        {\omega'}^{\mathsf T}(D_a-D_b)=0,\ 
        {\omega'}^{\mathsf T}(\delta_a-\delta_b)=0
    \right\}\\
    &\quad=
    m_T-
    \operatorname{rank}
    \begin{bmatrix}
        D_a-D_b & \delta_a-\delta_b
    \end{bmatrix}\\
    &\quad\leq m_T-1.
\end{aligned}
\]
A finite union of proper linear subspaces does not cover
$\R^{m_T}$. Choose
\[
    \omega\in
    \R^{m_T}\setminus
    \bigcup_{1\leq a<b\leq\iota}
    \left\{
        \omega'\in\R^{m_T}:
        {\omega'}^{\mathsf T}(D_a-D_b)=0,\ 
        {\omega'}^{\mathsf T}(\delta_a-\delta_b)=0
    \right\}.
\]
Then
\[
\begin{aligned}
    (\omega^{\mathsf T}D_a,\omega^{\mathsf T}\delta_a)
    &=
    (\omega^{\mathsf T}D_b,\omega^{\mathsf T}\delta_b)\\
    &\Longrightarrow a=b.
\end{aligned}
\]
Define
\[
    g(x):=\omega^{\mathsf T}\mathbf h(x),
\]
using the affine output layer with bias zero.

Let $U\subseteq\R^d$ be a nonempty connected open set on
which $g$ agrees with an affine map $L_U:\R^d\to\R$.
Since
$\bigcup_{P\in\mathcal P}\operatorname{int}(P)$
is dense in $\R^d$, choose $P_0\in\mathcal P$ satisfying
\[
    U\cap\operatorname{int}(P_0)\neq\varnothing,
    \quad
    a_0:=a(P_0).
\]
For every $P\in\mathcal P$ with
$U\cap\operatorname{int}(P)\neq\varnothing$, and every
$x\in U\cap\operatorname{int}(P)$,
\[
\begin{aligned}
    \omega^{\mathsf T}D_{a(P)}
    &=
    (\nabla L_U)^{\mathsf T},\\
    \omega^{\mathsf T}\delta_{a(P)}
    &=
    L_U(x)-\omega^{\mathsf T}D_{a(P)}x\\
    &=
    L_U(x)-(\nabla L_U)^{\mathsf T}x\\
    &=
    L_U(0).
\end{aligned}
\]
Applying these identities to $P_0$ gives
\[
\begin{aligned}
    (\omega^{\mathsf T}D_{a(P)},
     \omega^{\mathsf T}\delta_{a(P)})
    &=
    (\omega^{\mathsf T}D_{a_0},
     \omega^{\mathsf T}\delta_{a_0})\\
    &\Longrightarrow a(P)=a_0.
\end{aligned}
\]
Hence
\[
    \mathbf h(x)=D_{a_0}x+\delta_{a_0},
    \quad
    x\in
    U\cap
    \bigcup_{P\in\mathcal P}\operatorname{int}(P).
\]
This set is dense in $U$. Continuity therefore yields
\[
    \mathbf h(x)=D_{a_0}x+\delta_{a_0},
    \quad x\in U.
\]

Thus every nonempty connected open affine set of $g$ is
also an affine set of $\mathbf h$. Conversely, an affine
set of $\mathbf h$ is an affine set of
$g=\omega^{\mathsf T}\mathbf h$.
Their maximal connected open affine regions coincide, so
\[
    \mathcal R(g)=\mathcal R(\mathbf h).
\]
The affine output uses the already counted $m_T+1$
parameter slots. Hence $g$ satisfies the budget and
\[
    \mathcal R^\star_{\mathrm{ReLU}}(d,T,\Lambda)
    \geq
    \mathcal R(g)
    =
    \mathcal R(\mathbf h).
\]

For $0\leq j\leq d-1$, since $m_T\geq d$,
\[
\begin{aligned}
    \frac{m_T-j}{d-j}-\frac{m_T}{d}
    &=
    \frac{d(m_T-j)-m_T(d-j)}{d(d-j)}\\
    &=
    \frac{j(m_T-d)}{d(d-j)}\\
    &\geq0.
\end{aligned}
\]
It follows that
\[
\begin{aligned}
    A_d(m_T,1)
    &=
    \sum_{j=0}^{d}\binom{m_T}j\\
    &\geq
    \binom{m_T}{d}\\
    &=
    \prod_{j=0}^{d-1}\frac{m_T-j}{d-j}\\
    &\geq
    \prod_{j=0}^{d-1}\frac{m_T}{d}\\
    &=
    \left(\frac{m_T}{d}\right)^d\\
    &\geq
    \left\lfloor\frac{m_T}{d}\right\rfloor^d.
\end{aligned}
\]
Moreover, $\mu_\Lambda\geq2d$ gives
\[
\begin{aligned}
    \left\lfloor\frac{\mu_\Lambda}{d}\right\rfloor
    &\geq
    \frac{\mu_\Lambda}{d}-1\\
    &=
    \frac{\mu_\Lambda}{2d}
    +
    \left(\frac{\mu_\Lambda}{2d}-1\right)\\
    &\geq
    \frac{\mu_\Lambda}{2d}.
\end{aligned}
\]
There are exactly $\nu_T$ odd-indexed hidden layers.
Therefore,
\[
\begin{aligned}
    \mathcal R^\star_{\mathrm{ReLU}}(d,T,\Lambda)
    &\geq
    \left[
        \prod_{\ell=1}^{T-1}
        \left\lfloor\frac{m_\ell}{d}\right\rfloor^d
    \right]
    A_d(m_T,1)\\
    &\geq
    \prod_{\ell=1}^{T}
    \left\lfloor\frac{m_\ell}{d}\right\rfloor^d\\
    &=
    \left\lfloor
        \frac{\mu_\Lambda}{d}
    \right\rfloor^{d\nu_T}\\
    &\geq
    \left(
        \frac{\mu_\Lambda}{2d}
    \right)^{d\nu_T}\\
    &\geq
    \left(
        \frac{\Lambda}{4d\eta_{d,T}}
    \right)^{d\nu_T}.
\end{aligned}
\]
Combining this with the ReLU upper bound yields
\[
    \mathcal R^\star_{\mathrm{ReLU}}(d,T,\Lambda)
    =
    \Theta_{d,T}(\Lambda^{d\nu_T}).
\]

For sufficiently large $\Lambda$, both feasible classes are
nonempty and their attainable region counts are bounded
positive integers. Thus the maxima in their definitions exist.

Finally, for
\[
    \Lambda\geq
    \max\{4e_{\mathrm{base}},4d\eta_{d,T}\},
\]
the established bounds imply
\[
\begin{aligned}
    \frac{
        \mathcal R^\star_{\mathrm{KAN}}(d,T,\Lambda)
    }{
        \mathcal R^\star_{\mathrm{ReLU}}(d,T,\Lambda)
    }
    &\geq
    \frac{
        \left(\Lambda/(4e_{\mathrm{base}})\right)^{dT}
    }{
        \chi_d^T\Lambda^{d\nu_T}
    }\\
    &=
    \frac{
        \Lambda^{d(T-\nu_T)}
    }{
        \chi_d^T(4e_{\mathrm{base}})^{dT}
    },
\end{aligned}
\]
and
\[
\begin{aligned}
    \frac{
        \mathcal R^\star_{\mathrm{KAN}}(d,T,\Lambda)
    }{
        \mathcal R^\star_{\mathrm{ReLU}}(d,T,\Lambda)
    }
    &\leq
    \frac{
        \chi_d^T\left(\Lambda/(2T)\right)^{dT}
    }{
        \left(\Lambda/(4d\eta_{d,T})\right)^{d\nu_T}
    }\\
    &=
    \frac{
        \chi_d^T(4d\eta_{d,T})^{d\nu_T}
    }{
        (2T)^{dT}
    }
    \Lambda^{d(T-\nu_T)}.
\end{aligned}
\]
All multiplicative constants depend only on $d$ and $T$.
Since
\[
    T-\nu_T
    =
    T-\left\lceil\frac T2\right\rceil
    =
    \left\lfloor\frac T2\right\rfloor,
\]
we conclude that
\[
    \frac{
        \mathcal R^\star_{\mathrm{KAN}}(d,T,\Lambda)
    }{
        \mathcal R^\star_{\mathrm{ReLU}}(d,T,\Lambda)
    }
    =
    \Theta_{d,T}\!\left(
        \Lambda^{d\lfloor T/2\rfloor}
    \right).
\]
This completes the proof.
\end{proof}

\end{document}